%% file: cvpr.tex
\documentclass[final]{cvpr}

\usepackage{times}
\usepackage{epsfig}
\usepackage{graphicx}
\usepackage{amsmath}
\usepackage{amssymb}
\usepackage{algorithm}
\usepackage{algorithmic}
\usepackage{booktabs}
\usepackage{multirow}

\usepackage[pagebackref=true,breaklinks=true,colorlinks,bookmarks=false]{hyperref}

\def\cvprPaperID{19} 
\def\confYear{CVPR 2021}

\begin{document}

\title{Boosting Adversarial Robustness using Feature Level Stochastic Smoothing}

\author{Sravanti Addepalli\thanks{Equal contribution. \newline Correspondence to: Sravanti Addepalli (sravantia@iisc.ac.in), Samyak Jain (samyakjain.cse18@itbhu.ac.in)}~, ~Samyak Jain\footnotemark[1]~, ~Gaurang Sriramanan\footnotemark[1]~, ~R. Venkatesh Babu\\
Video Analytics Lab, Department of Computational and Data Sciences\\ Indian Institute of Science, Bangalore, India\\}

\maketitle

\begin{abstract}
Advances in adversarial defenses have led to a significant improvement in the robustness of Deep Neural Networks. However, the robust accuracy of present state-of-the-art defenses is far from the requirements in critical applications such as robotics and autonomous navigation systems. Further, in practical use cases, network prediction alone might not suffice, and assignment of a confidence value for the prediction can prove crucial. In this work, we propose a generic method for introducing stochasticity in the network predictions, and utilize this for smoothing decision boundaries and rejecting low confidence predictions, thereby boosting the robustness on accepted samples. The proposed Feature Level Stochastic Smoothing based classification also results in a boost in robustness without rejection over existing adversarial training methods. Finally, we combine the proposed method with adversarial detection methods, to achieve the benefits of both approaches.
\end{abstract}
\vspace{-0.3cm}
\section{Introduction}
\label{sec:intro}
Deep Neural Networks are susceptible to carefully crafted imperceptible noise known as adversarial attacks \cite{goodfellow2014explaining}, which can flip their predictions to completely unrelated classes with high confidence. The catastrophic impact of such attacks has led to significant interest towards building defenses against such attacks. 

Adversarial training using Projected Gradient Descent (PGD), proposed by Madry \etal in 2018 \cite{madry-iclr-2018} has been one of the most successful defenses so far. PGD adversarial training coupled with early stopping is still one of the leading defenses \cite{rice2020overfitting, pang2020bag} indicating that progress on the front of adversarial defenses has been meager since $2018$. Schmidt \etal \cite{schmidt2018adversarially} show that adversarial training requires significantly more data when compared to standard training. On the CIFAR-10 dataset for example, Carmon \etal \cite{carmon2019unlabeled} demonstrate that a $7\%$ increase in robustness requires $500$K samples in addition to the $50$K training samples of the original dataset. However, it is not practical to assume the availability of $10\times$ more data for training robust models.

Another avenue of research has been towards detecting adversarial samples \cite{yin2019adversarial, roth2019odds, hu2019new, ghosh2019resisting}. Such methods can be used to detect whether a test sample is adversarial or not, allowing the system to abstain from prediction on adversarial samples. However, most of the detection methods rely on identifying specific properties of adversarial images. Hence, they work well under a black-box setting, but fail in the presence of an adaptive adversary, whose goal is to craft an adversarial example which is similar to the distribution of natural images, specifically with respect to the property that is used for detection \cite{tramer2020adaptive}. 

Recent work by Stutz \etal \cite{stutz2019confidence} demonstrates a method of detecting different types of adversaries using Confidence Calibrated Adversarial Training (CCAT). The authors propose to train networks that assign very low confidence to adversarial samples. This method is shown to induce low-confidence predictions to adversaries constrained in other $\ell_p$-norm balls as well, and demonstrates remarkably high detection accuracies, while limiting the rejection rate on correctly classified clean samples to $1\%$.  However, this method is overly sensitive to even small random perturbations, which could occur due to factors such as slight soiling on a sensor collecting data. Based on our evaluation on the CIFAR-10 dataset, CCAT rejects $78.60\%$ of images corrupted with Bernoulli noise of magnitude $1/255$, while a model with standard training has an accuracy of $92.4\%$ on the same (Ref: Table-\ref{table:ccat_noise} in the Supplementary). Thus, the model is becoming more sensitive, while the requirement in general is to make models more robust. Additionally, as shown in Fig.\ref{fig:ccat}(a) in the Supplementary, CCAT incorrectly accepts a large fraction ($36.5\%$) of adversarial examples at low perturbation magnitudes ($\delta=3/255$) against an adaptive attack proposed by the authors \cite{stutz2019confidence}. 

Adversarial detection alone cannot meet the requirements of applications such as robotics and autonomous navigation systems. As an example, at fast driving speeds, a self-driving car abstaining from prediction can be dangerous, as the driver may not be able to take control instantly. Hence we need to limit the frequency at which the system abstains from prediction, while also being able to reject hard images. 
An ideal system should therefore operate such that weak adversaries which can be correctly classified by an adversarially trained model are not rejected, while strong adversaries which are likely to be misclassified are rejected. Towards this end, we propose a unified framework that combines the merits of adversarial training and detection, while also overcoming the shortcomings of both. 

\vspace{0.1cm}
\noindent We list the key contributions in our paper here:
\vspace{-0.1cm}
\begin{itemize}
\itemsep0em 
    \item We propose Adversarial training using a stochastic classifier, which enforces one of the feature layers to follow a predefined distribution, facilitating sampling from the same during training and deployment.
    \item We propose Feature Level Stochastic Smoothing to achieve a boost in adversarial robustness over standard deterministic classifiers. 
    \item We use the proposed stochastic classifier for rejecting low confidence samples, thereby resulting in a significant boost in adversarial robustness. 
    \item We propose metrics to evaluate classifiers which improve robustness using adversarial training, while also incorporating a rejection scheme. 
    \item Finally, we propose a scheme of combining our approach with Confidence-Calibrated Adversarial Training (CCAT) \cite{stutz2019confidence}, to achieve the merits of both. 
\end{itemize}

\noindent The code and pretrained models are available at: \\ \url{https://github.com/val-iisc/FLSS}

\section{Related Works}
\textbf{Types of Adversarial Defenses:} Amongst the most common methods used to produce robust networks is Adversarial Training, wherein the network is exposed to adversarial images during the training regime. 
Several other methods relied on introducing randomized or non-differentiable components either in the pre-processing stage or in the network architecture, so as to minimise the effectiveness of generated gradients. However, Athalye \etal \cite{athalye2018obfuscated} broke several such defense techniques \cite{buckman2018thermometer,ma2018characterizing,s.2018stochastic,xie2018mitigating,song2018pixeldefend}, where it was shown that methods which relied on gradient obfuscation were not truly robust, as gradient masking effects could be successfully circumvented by an adaptive adversary. Though our method uses stochastic elements, the network remains completely differentiable end-to-end through the reparamaterisation trick, akin to that used in Variational Autoencoders \cite{kingma2013auto}, and  does not rely on gradient obfuscation to achieve robustness. Further, we present a thorough evaluation of our model based on attacks introduced in \cite{athalye2018obfuscated} to verify the same.

\textbf{Adversarial Training based Defenses:} Madry \etal \cite{madry-iclr-2018} proposed training on multi-step adversaries generated using Projected Gradient Descent (PGD), so as to minimise the worst-case loss within the given constraint set. PGD based training continues to be one of the most effective defenses against adversarial attacks known till date. He \etal \cite{he2019parametric} proposed to inject trainable Gaussian noise in the weights of the network as a regularization method while performing standard PGD training. This transforms the weight tensor to a noisy one, wherein the variance of the added Gaussian is parameterized and trainable. Further, Zhang \etal \cite{zhang2019theoretically} introduced a framework to trade-off clean accuracy for adversarial robustness, using a multi-step training method called TRADES. Though TRADES was shown be more effective than PGD training, recent works such as that of Rice \etal  \cite{rice2020overfitting} indicate that high capacity models trained using PGD with early stopping can achieve similar, if not better results. Following this, Pang \etal  \cite{pang2020bag} showed that by tuning parameters such as weight decay and the step learning rate schedule, TRADES can achieve better robust accuracy compared to that of Rice \etal. The current state-of-the-art robust accuracy is achieved using Adversarial Weight Perturbation (AWP) \cite{wu2020adversarial} where the loss maximization is done with respect to both input pixels and weight space of the network. Minimizing loss on a proxy network with perturbed weights is shown to result in significantly improved generalization to the test set, owing to the improved flatness of loss landscape. While the AWP formulation can be combined with any defense, AWP-TRADES achieves the state-of-the-art results currently.

\textbf{Detection of Adversarial Examples:} Another avenue of research in this field has been towards addressing the detection of adversarial perturbations. 
Gosh \etal \cite{ghosh2019resisting} use a VAE with Gaussian Mixture density to perform thresholding based on the distance between the encoding of an input sample and the encoding of the predicted class label, combined with thresholding of the reconstruction error obtained from the decoder. However, this method is not scalable to datasets such as CIFAR-100, with a large number of classes. 

Several other detection methods exist which seek to exploit the differences between clean and adversarial samples with respect to a given property. However, such methods are often effective only in black-box settings, and are susceptible to adaptive attacks in a white-box setting. Tramer \etal \cite{tramer2020adaptive} systematically evaluate several defense methods, comprising both adversarial training as well as detection methods. The authors show that with carefully crafted adaptive attacks, several detection methods \cite{roth2019odds,yin2019adversarial, hu2019new} could be circumvented. The detection method by Roth \etal \cite{roth2019odds} relies on  the fact that network outputs of adversarial samples exhibit higher sensitivity to input noise when compared to natural images. The method incorporates thresholding of logits, which inherently assumes that adversarially perturbed samples either have highly confident scores for the incorrect class or lie abnormally close to decision boundaries, depending on the nature of the attack used (such as PGD \cite{madry-iclr-2018} or CW \cite{carlini2017towards} attacks). This detection method could be compromised using a feature level attack \cite{sabour2015adversarial}, which attempts to generate adversarial examples that exhibit properties of natural images by utilizing a guide image from a different class. We show that our method is robust against this class of feature level adversaries with different loss functions as well (Ref: Table-\ref{table:cifar_adaptive} in the Supplementary).

\textbf{Randomized Smoothing (RS): }Cohen \etal \cite{cohen2019certified} proposed the addition of Gaussian noise to input images during inference, in order to generate models that are certifiably robust to perturbations that lie within a specified $\ell_2$ norm bound. The addition of Gaussian random noise can be used to produce a distribution of network predictions; a p-test on the top two most frequently predicted classes can then be utilised to identify the confidence level of the final averaged prediction. A given image can be rejected if its averaged prediction confidence lies below a pre-set threshold. In this work, we propose to add Gaussian noise in feature space and train the model to produce consistent predictions across multiple samples drawn. We obtain improved performance when compared to randomized smoothing baselines.

\section{Preliminaries}
\label{sub:threat}
In this paper, we consider a stochastic classifier $C_{\theta}$, that maps an input image $x$ to its corresponding softmax output $C_{\theta}(x,\epsilon)$ after sampling a noise vector $\epsilon$ from a fixed probability distribution such as the Standard Normal distribution $\mathcal{N}(0,I)$ with zero mean, and Identity covariance matrix. The stochastic classifier has two primary components: an Encoder network $E= \{ \mu_{\theta_{E}}, \Sigma_{\theta_E}\}$, and a Multi-layer Perceptron $M$ (consisting of one or more layers), such that:

$C_{\theta}(x,\epsilon) = M(\mu_{\theta_{E}}(x) +  \Sigma_{\theta_E}(x) \odot \epsilon )$

\noindent where $\odot$ denotes element-wise multiplication. For an image $x$, we denote its corresponding ground-truth label as $y$. For a given sampled noise vector $\epsilon$, we denote the cross-entropy loss for a data sample $\{x, y\}$ as $\ell_{CE}(C_{\theta}(x,\epsilon),y)$. Further, given a clean image $x$, we denote an adversarially modified counterpart as $\widetilde{x}$.
In this paper, we primarily consider adversaries that are constrained in $\ell_{\infty}$ norm of $\delta=8/255$:

$\mathcal{A}(x)=\{\widetilde{x}: ||x-\widetilde{x}||_\infty \leq \delta \} $ 

\begin{figure}
\centering
        \includegraphics[width=\linewidth]{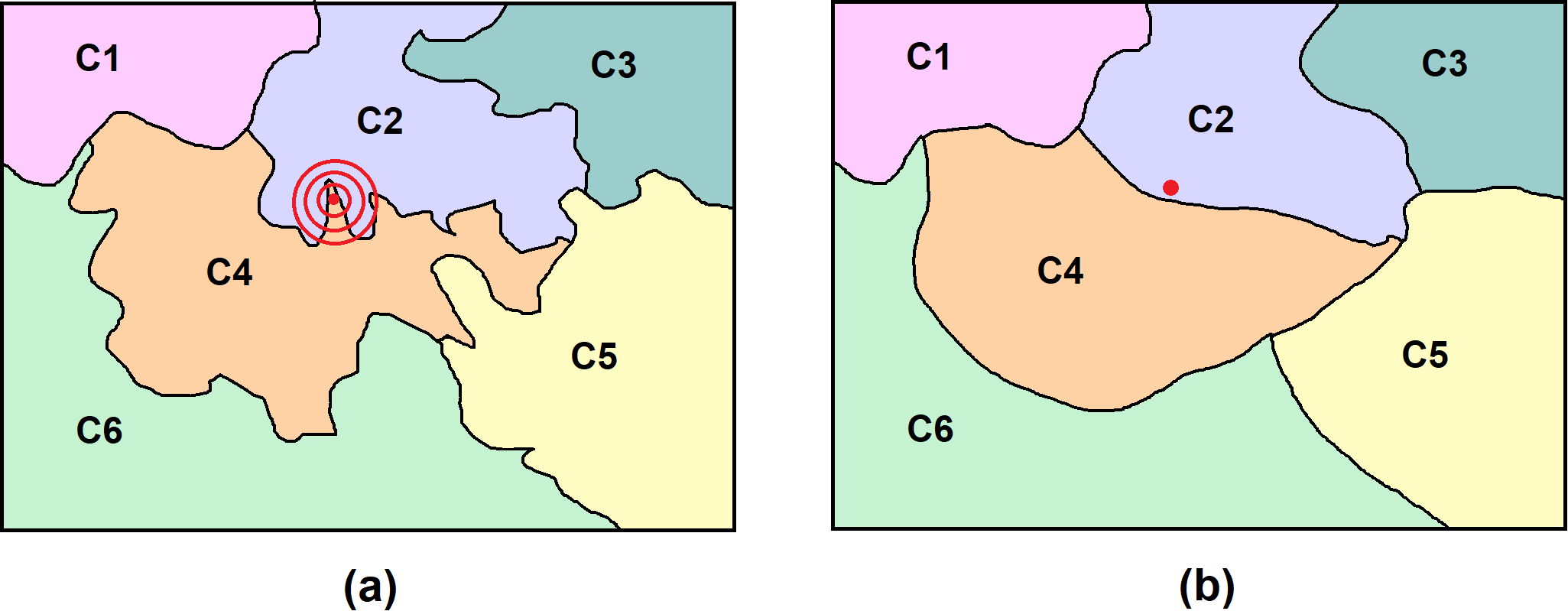}
        \vspace{-0.5cm}
        \caption{Decision boundaries (in feature space) of (a) Standard Classifier and (b) Feature Level Stochastic Smoothing  based Classifier. The data sample which belongs to the class C2 gets incorrectly predicted as C4 in (a), whereas in (b) it is predicted correctly as C2. The smoothed classifier considers a majority vote over samples within the local neighborhood of the image as shown in (a).}
        \label{fig:smoothing}
\end{figure}

\begin{figure}
\centering
        \includegraphics[width=\linewidth]{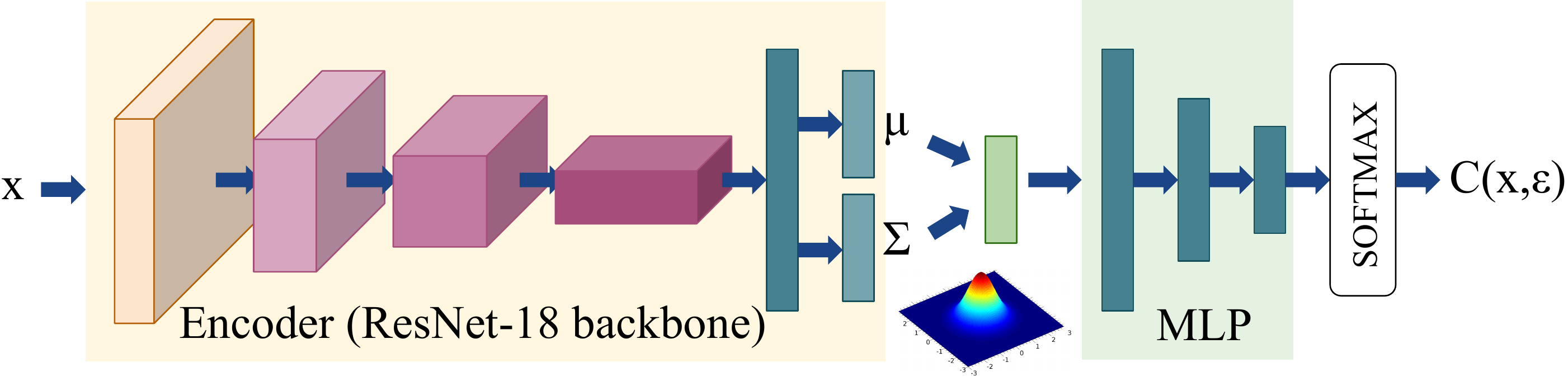}
        \vspace{-0.5cm}
        \caption{Feature Level Stochastic Smoothing Classifier: The network is trained such that output of the encoder follows a fixed distribution, thereby enabling sampling from this layer to generate a randomized pool of network outputs for a given input $x$.}
        \label{fig:network-architecture}
\end{figure}
\section{Proposed Approach}

In this section, we discuss the proposed approach in detail. We first discuss the proposed stochastic smoothing based classifier, followed by details on how such classifiers can be used to boost adversarial robustness. 

\subsection{Feature Level Stochastic Smoothing}
\label{subsec:FLSS}
Standard classifiers are deterministic and can be defined as a function mapping from input space to output space, leading to a unique output for every possible input in the domain of the function. Deep Networks can be used as deterministic classifiers and are known to achieve very high classification performance. However, training Deep Networks using the standard cross-entropy loss is known to force the network to predict outputs with a high confidence, even for out-of-distribution samples. Several methods \cite{zhang2017mixup,stutz2019confidence,pereyra2017regularizing} have been proposed to overcome this, allowing the confidence predictions of the classifier to be more informative. Some of these methods are also known to generate smoother decision boundaries \cite{zhang2017mixup}.

In contrast to deterministic classifiers, some classifiers introduce stochasticity during training, typically for their regularizing effect. Randomness during test time is not preferred, as many applications require consistency in predictions while being deployed. This is avoided by taking expectation over the random components during test time either analytically, as done in dropout \cite{srivastava2014dropout}, or by finding a sample mean using a sufficiently large number of samples during test time, as done in randomized smoothing \cite{cohen2019certified}. In this work, we propose Feature Level Stochastic Smoothing based classification, which enforces one of the feature layers to follow a predefined distribution such as the standard normal distribution, thereby facilitating sampling from the same during training and deployment. We use this to achieve the dual objective of smoothing decision boundaries, and for generating multiple randomized predictions for rejection. The diagram in Fig.\ref{fig:smoothing} shows that feature space decision boundaries of standard classifiers tend to be highly non-smooth, primarily due to the high dimensionality of the network parameters, leading to poor generalization, specifically to out-of-distribution samples. On the other hand, in the proposed method, we generate predictions by taking an expectation over randomized feature vectors, leading to smoother decision boundaries and better generalization. The expectation is approximated using a sample mean, and consistency among all outputs is used to estimate confidence of the prediction and decide whether the input sample should be rejected.  

An implementation of the proposed Feature Level Stochastic Smoothing Classifier is shown in Fig.\ref{fig:network-architecture}. Motivated by the Variational Autoencoder formulation \cite{kingma2013auto}, we enforce the encoder network to predict mean ($\mu_{\theta_{E}}$) and variance ($\Sigma_{\theta_E}$) vectors for each input sample, and use the reparameterization trick for ease of backpropagation. Hence, rather than sampling directly from the mean and variance predicted by the network, we sample from the standard normal distribution and pass $\mu_{\theta_{E}} + \Sigma_{\theta_E} \odot \epsilon$ to the rest of the network. 
The sampled feature vector is passed as input to the multi-layer perceptron head for the final classification. Every sampled feature vector gives a different random prediction. We describe the algorithm used for training and inference during test time in the following sections. 

\subsection{Training Algorithm}

The proposed method uses feature level stochastic smoothing discussed in the above section for training robust classifiers. 
The algorithm for training is shown in Algorithm-\ref{alg:train_algo}. For ease of notation, we consider a single sample at a time in the algorithm, however the training is done using mini-batches of size 128. In every mini-batch, two steps of training are done. The first step is meant for training on adversarial samples, while the second step uses only clean samples for training. We use adversarial samples constrained within an $\ell_\infty$ norm of $\delta=8/255$ for training. An adversarial sample is generated using a $10$-step PGD attack \cite{madry-iclr-2018} and a $11^{th}$ AWP step \cite{wu2020adversarial}. An adversarial sample is initialized by combining an input image with noise sampled from $\mathcal{U}(-\delta,\delta)$, and is passed through the encoder $E$. A random vector, $\epsilon$ is sampled from the Standard Normal distribution and the reparameterized vector, $\mu_{\theta_{E}} + \Sigma_{\theta_E} \odot \epsilon$ is passed on to the MLP. Cross entropy loss is computed at the output of the network, and is maximized over $10$ iterations to find the perturbed image. Following this, a $11^{th}$ iteration is used to perturb the network weights to maximize the training loss in Eq.\ref{eq:step1} using AWP \cite{wu2020adversarial}. During all $11$ iterations, the same initial sampled $\epsilon$ value is used in order to ensure that the maximization objective remains consistent throughout. The adversary corresponding to the image $x_i$ is denoted by $\widetilde{x}_i$ and is used in the following loss which is minimized during the first step of training:
\begin{equation}
\label{eq:step1}
     L = \ell_{CE}(C_{\theta} (\widetilde{x}_i,\epsilon) , y_i ) + KL_1 + KL_2 + KL_3
\end{equation}

The first loss term corresponds to the cross-entropy loss on adversarial samples, as utilised in PGD adversarial training \cite{madry-iclr-2018} to obtain robust models. As shown in Algorithm-\ref{alg:train_algo} (L-6), $KL_1$ denotes the Kullback-Leibler (KL) divergence between the Gaussian distribution $\mathcal{N}(E(x_i))$ corresponding to the clean sample and the Standard Normal distribution. This term is adapted from the Variational Autoencoder \cite{kingma2013auto} setting, and is crucial to enforce the feature representations to follow a known distribution, which in this case is a Standard Normal distribution. The $KL_2$ term (L-7 in Algo.\ref{alg:train_algo}) is the KL divergence between the Gaussian distributions corresponding to a pair of adversarial and clean samples. This aids the encoder to learn a smooth function mapping in the $\ell_{\infty}$ ball of radius $\delta$ around each image, thereby assisting the adversarial training of the network. We use the closed form expression of KL divergence between two Gaussian distributions for the realization of $KL_1$ and $KL_2$. We additionally minimize $KL_3$ (L-8 in Algo.\ref{alg:train_algo}), which is the KL divergence between the softmax outputs of an adversarial image with and without sampling. This encourages the network to produce consistent predictions across various samples of an adversarial image.

In the second step, the network is trained on clean samples using the following losses:

\begin{equation}
\label{eq:step2}
     L = \ell_{CE}(C_{\theta} (x_i,\epsilon') , y_i ) + KL_1 + KL_4 
\end{equation}

Here, the first term denotes the cross-entropy loss on clean samples, and the second term, $KL_1$ denotes the objective of enforcing the output of encoder to follow a Standard Normal distribution, as seen earlier. The third term, $KL_4$ (L-13 in Algo.\ref{alg:train_algo}), is the KL divergence between the softmax predictions of a clean image and a sampled clean image. This loss is crucial to ensure consistency in predictions across various samples of a given clean image, thereby leading to improved non-rejection of clean samples. We present ablation experiments in Table-\ref{table:ablations} in the Supplementary, to highlight the significance of each of the loss terms.
\input{Training_Algo}

\subsection{Rejection Scheme}
\label{sub-sec:rejection}
In addition to the goal of improving adversarial robustness, the proposed method also rejects samples which are hard to classify, thereby leading to a boost in the accuracy of accepted samples. In this section, we discuss details on the rejection scheme proposed to be used during deployment of the classifier. Every test sample would be passed through the Encoder network in order to obtain its corresponding mean and variance vectors. At the output of the encoder, N vectors are sampled from the Gaussian corresponding to these mean and variance vectors, and are further propagated through the MLP network to obtain N softmax vectors.  

As discussed in Section-\ref{subsec:FLSS}, introduction of stochasticity during training is generally coupled with the use of expected value during test time, in order to obtain regularized and deterministic outputs. In the absence of an analytical expression for the expectation, a sample mean over all N probability vectors can be used. It is to be noted that, as N approaches infinity, the sample mean would approach the expected value. Thus, higher values of N lead to much better estimates of the network output as shown in Fig.\ref{fig:fpr}(a) in the Supplementary, while they add to the test time complexity.  
We choose N to be 100 for our experiments. 
Since the N additional forward propagations are done only on the MLP, the test time overhead is insignificant for $N=100$. We note that the increase is computational cost is $2\%$ with $100$ times sampling when parallelized on Nvidia-2080Ti.

In the proposed rejection scheme, we first find the class predictions for each of the N sampled outputs. Further, we define the class with maximum number of predictions (majority vote class) to be the class predicted by the \textit{Smoothed Classifier}. We set rejection threshold based on frequency of the predicted class, which serves as a proxy to the confidence of prediction. If the frequency of the predicted class is below a predefined threshold $f$, the classifier rejects the sample, otherwise it returns the most frequent class as its prediction. We empirically find that the majority vote based rejection scheme leads to better improvements in robust accuracy, when compared to a rejection scheme based on finding sample mean across softmax predictions.

We discuss the important metrics for our proposed classifier along with the method used for selecting threshold in the following section.

\subsection{Evaluation Metrics}
\label{sub:eval metrics}
The commonly used evaluation metrics for adversarially trained classifiers are accuracy on clean (or natural) samples, $Acc_{nat}$ and accuracy on adversarial samples, $Acc_{adv}$. For methods which detect adversarial samples, the important metrics include True Positive Rate (TPR) and False Positive Rate (FPR), where the case of rejection is set to be the positive class. While it is important to ensure that all adversaries are detected (TPR), it is also necessary to limit the number of clean samples which are incorrectly predicted as adversarial, and hence rejected (FPR). The proposed classifier combines both adversarial training as well as detection, and therefore requires novel metrics which can better measure the effectiveness of the method. 

\textbf{Selection of threshold for rejection: }We select the threshold for rejection such that not more than $10\%$ of the clean samples are correctly classified and rejected \cite{stutz2019confidence}. It is to be noted that this metric is independent of the clean accuracy of the classifier, and hence the number of correctly classified clean samples that are allowed to be rejected are the same across all baselines for a given dataset. In practice, a hold-out validation set can be used for finding this threshold. However, in order to strictly ensure a fair comparison between baselines and the proposed method, we use the test set to find the threshold. 

\textbf{Metrics used for evaluation: } We explain the terminology and metrics used for our evaluations here. We denote the accuracy on natural samples and adversarial samples in the \textit{No Sampling} ($NS$) case by $Acc_{nat,NS}$ and $Acc_{adv,NS}$ respectively. In the proposed classifier, this metric is calculated by passing the mean vector from the encoder output directly to the MLP, without considering the variance. $Acc_{nat,0\%}$ and $Acc_{adv,0\%}$ denote the accuracy without rejection (but with sampling) on natural samples and adversarial samples respectively, while $Acc_{nat,10\%}$ and $Acc_{adv,10\%}$ denote the same with the rejection threshold set to $10\%$. For defining $Acc_{adv,10\%}$, we consider the worst case attack for every data sample as recommended by Carlini \etal \cite{carlini2019evaluating}. The calculation of this metric along with other important metrics is described below.

\textbf{Worst case robustness evaluation with rejection: } For a data sample $\{x_i,y_i\}$, we denote the predicted label using any given classifier $C$ by $C(x_i)$ and the decision of the rejection scheme (detector) by $D(x_i)$. $D(x_i)=1$ denotes the case where the sample is rejected, while $D(x_i)=0$ means that the sample is accepted by the classifier for prediction. 

We denote the set of all perturbations of a data sample $x_i$, within the threat model defined in Section-\ref{sub:threat}, by $\mathcal{A}(x_i)$. 

We define $S_{FC}$ (FC: Flag Correct) to be the set of all images which are not rejected by any adversary, and are predicted correctly by the classifier as shown below: 
\begin{equation}
    S_{FC} = \{i:D(\widetilde{x}_i)=0~, C(\widetilde{x}_i)=y_i ~~ \forall~ \widetilde{x}_i \in \mathcal{A}(x_i)\}
\end{equation}
$S_{FW}$ (FW: Flag Wrong) is defined as the set of all accepted images, incorrectly predicted for at least one attack:
\begin{equation}
    S_{FW} = \{i:~\exists ~\widetilde{x}_i \in \mathcal{A}(x_i) : D(\widetilde{x}_i)=0, C(\widetilde{x}_i)\neq y_i\}
\end{equation}
$S_{FC}$ is computed in practice by obtaining the indices of correctly classified  accepted samples for each attack, and finding an intersection of all such sets. Similarly, $S_{FW}$ is computed by finding the indices of incorrectly classified samples which are accepted for each attack, and subsequently finding a union across all such sets. 
\begin{equation}
    S_{FC} = {\bigcap}_{i} S_{FC,attack_i}~,~~ S_{FW} = {\bigcup}_{i} S_{FW,attack_i}
\end{equation}
We define the metrics FC and FW as the percentage of images that belong to the sets $S_{FC}$ and $S_{FW}$ respectively as follows, where $\mathcal{X}$ denotes the test set: 
\begin{equation}
    FC = \frac{|S_{FC}|}{|\mathcal{X}|} \cdot 100~,~~FW = \frac{|S_{FW}|}{|\mathcal{X}|} \cdot 100~
\end{equation}

We define the worst case robust accuracy on accepted samples to be the fraction of samples which are always accepted and correctly classified, under any possible adversarial attack. This fraction is defined on the set of images which are either always correctly classified or are incorrectly predicted by at least a single attack. 

\begin{equation}
    Acc_{adv,10\%} = \frac{|S_{FC}|}{|S_{FC}|+|S_{FW}|} \cdot 100
\end{equation}

We define $R$ to be the set of all images which can be rejected using at least a single attack. $MPR$ (Maximum Percentage Rejection) is defined as the maximum percentage of samples that can be rejected by the model. 
\begin{equation}
    R = \{i: \exists~~\widetilde{x}_i \in \mathcal{A}(x_i) : D(\widetilde{x}_i)=1\},~ MPR = \frac{|R|}{|\mathcal{X}|}\cdot100
\end{equation}
For adversarial detection methods, $MPR$ would be equal to $100\%$ in the ideal case, as the goal of such methods is to reject all adversarial samples. 
This makes them susceptible to adversaries who aim to get all input images rejected. 

\section{Experiments and Results}

We report results on the standard benchmark datasets, CIFAR-10 and CIFAR-100 \cite{krizhevsky2009learning}. We use ResNet-$18$ architecture \cite{he2016deep} for all the experiments. We report all results with an $\ell_{\infty}$ constraint of $8/255$. We present more details on datasets, training settings and results in the Supplementary.

\input{tables/main_all}
\subsection{Baselines}
In this paper, we approach the problem of improving the adversarial robustness of Deep Networks and rejecting low confidence samples in parallel. There is no prior work which we can directly compare our results with. Stutz \etal \cite{stutz2019confidence} show that confidence thresholding of state-of-the-art models such as PGD \cite{madry-iclr-2018} and TRADES \cite{zhang2019theoretically} give the best results for detection as well, when limited to a well-defined threat model. Therefore, PGD, TRADES and AWP \cite{wu2020adversarial} trained models can be used to achieve the combined goal of improving adversarial robustness, and also rejecting low confidence samples. We use these models combined with confidence thresholding as our primary baselines. If the confidence of the predicted class is lower than a predefined threshold, the samples are rejected, otherwise a classification output is predicted. The rejection threshold is set based on the same criteria as that defined in Section-\ref{sub:eval metrics}. 

In addition to the baselines discussed above, we also consider the baselines PGD (Noise) and TRADES (Noise). These baselines are same as PGD or TRADES during training, however they differ during test time. For every test image, we generate $100$ noise images by sampling each pixel from $\mathcal{U}[-32/255,32/255]$ distribution. Each of these attack images is added to the test image, to generate 100 samples of the test image. These $100$ samples are passed through the network to generate $100$ softmax vectors. We implement the same rejection scheme that is used in the proposed approach for generating the predictions and for rejecting low confidence samples. For this baseline, reducing noise results in a very low rejection percentage, whereas increasing noise reduces the accuracy on clean samples significantly. We add noise only during inference (and not training) since we empirically find that training such models with Gaussian Noise augmentations results in degraded performance. 

We consider the baseline of Randomized smoothing (RS) \cite{cohen2019certified}, where we train the model as described by the authors. We also consider a baseline which combines Randomized Smoothing with the TRADES defense, as proposed by Blum \etal \cite{blum2020random}, and reject images if the most frequently predicted class is less than a pre-defined threshold value. For the baseline using Parametric Noise Injection (PNI) \cite{he2019parametric}, we utilize the stochasticity of the model to find a threshold corresponding to the criteria of rejecting $10\%$ correctly classified natural images.

We further compare our method with the work by Stutz \etal \cite{stutz2019confidence}, although they primarily prove robustness to unseen threat models, while in this work, we consider the robustness within a well defined threat model. 

\subsection{Attacks considered for Evaluation}

The evaluations are done to predict the worst case accuracy across an ensemble of attacks. Across all evaluations (unless specified otherwise), we set the rejection threshold such that not more than $10\%$ of the clean samples are correctly classified and rejected. The metrics discussed in Section-\ref{sub:eval metrics} are reported.

Table-\ref{table:cifar_1} in the Supplementary shows results of the proposed approach and baseline models against an ensemble of five attacks: Projected Gradient Descent (PGD) with fixed step size, AutoPGD with Cross-Entropy loss (APGD-CE) and Difference of Logits Ratio Loss (APGD-DLR) \cite{croce2020reliable}, Fast Adaptive Boundary Attack (FAB) \cite{croce2019minimally} and Square Attack \cite{andriushchenko2019square}. The first four comprise of some of the strongest known white-box attacks, while Square attack is a query based black-box attack. We use 100 steps each for PGD, APGD-CE, APGD-DLR and FAB attack, and use 5000 queries for the Square attack. 
We observe that the proposed method achieves significantly better adversarial performance across different metrics. Based on these results, we select the following strong baselines for evaluations in the main paper: PGD, TRADES and AWP with confidence thresholding, and CCAT. 

For our main evaluations in Table-\ref{table:main_all}, we use the following ensemble of $100$-step attacks, which were able to reliably estimate the worst-case performance of networks before and after rejection, at a reasonable computational budget: PGD \cite{madry-iclr-2018}, APGD-CE, APGD-DLR \cite{croce2020reliable}, PGD with CW loss \cite{carlini2017towards}, GAMA-PGD and GAMA-MT \cite{sriramanan2020gama}. While AutoAttack \cite{croce2020reliable} is strong enough to estimate robustness before rejection, we find that Maximum-Margin based attacks such as PGD with CW loss, GAMA-PGD and GAMA-MT are significantly better at estimating the true robustness after rejection, possibly because rejection implicitly relies on the confidence-margin of predictions.  

\subsection{Results}

Test-time prediction using the proposed method (FLSS) involves sampling $100$ times from the latent space, after which the majority vote class is predicted. While we use a Standard Normal distribution $\mathcal{N}(0,I)$ for sampling during training, the use of different scaling factors for standard deviation at test time can result in different robustness-accuracy trade-offs. For example, as shown in the CIFAR-$10$ results in Table-\ref{table:main_all}, we achieve $2.1\%$ higher robust accuracy after rejection ($Acc_{adv,10\%}$) using standard deviation scaling of 2 ($SD=2$), when compared to 1. While the clean accuracy without rejection ($Acc_{nat,0\%}$) for $SD=2$ is $2.83\%$ lower than the case with $SD=1$, we achieve higher natural accuracy after rejection ($Acc_{nat,10\%}$) at $SD=2$, which is the main metric to consider. Therefore, we achieve better clean accuracy ($Acc_{nat,10\%}$) and adversarial robustness ($Acc_{adv,10\%}$) after rejection using $SD=2$. We empirically find that increasing the scaling factor does not improve performance further. Hence, we consider $SD=2$ as our primary approach across all datasets. 

Overall, the important evaluation metrics to consider are natural accuracy ($Acc_{nat,10\%}$)  and robust accuracy ($Acc_{adv,10\%}$) after rejection. We note that the proposed method FLSS ($SD=2$) achieves a significantly higher robust accuracy after rejection ($Acc_{adv,10\%}$)  when compared to the baselines across both datasets, at a comparable value of clean accuracy after rejection ($Acc_{nat,10\%}$). While the clean accuracy of CCAT \cite{stutz2019confidence} is exceptionally high ($97.52\%$), the results of CCAT cannot be directly compared with ours, as CCAT is an algorithm for detection of adversaries. We discuss CCAT in detail in Section-\ref{sec:ccat}. 

For CIFAR-10, we obtain an improvement of $3.15\%$ on the $Acc_{adv,10\%}$ metric over the strongest baseline (AWP \cite{wu2020adversarial}). For CIFAR-$100$ we achieve an improvement of $3.89\%$ over PGD-AT \cite{madry-iclr-2018,pang2020bag,rice2020overfitting}, which is the strongest baseline. We also obtain the best Clean Accuracy after rejection ($Acc_{nat,10\%}$) when compared to all baselines on CIFAR-$100$.

\subsection{Combining FLSS with CCAT}  
\label{sec:ccat}

As discussed in Section-\ref{sec:intro}, although CCAT \cite{stutz2019confidence} achieves remarkably high detection accuracy while limiting the rejection percentage on correctly classified clean samples to $1\%$, it is overly sensitive to adversarial examples and random perturbations at low magnitudes. In some cases it achieves a very high rejection rate on easy samples, and in others it causes high misclassification on accepted samples. 
While the proposed approach (FLSS) achieves a significantly higher accuracy on adversarial samples after rejection ($Acc_{adv,10\%}$), we limit the rejection percentage of correctly classified clean images to $10\%$, which is much higher than that considered in CCAT ($1\%$). Although Table-\ref{table:main_all} reports CCAT results with a threshold of $10\%$, the authors achieve the claimed results with a threshold of $1\%$. 

We propose to achieve the merits of CCAT and FLSS (Ours) by using a combination of both models during test-time. The test samples would first be evaluated using CCAT to obtain a decision of Accept or Reject. Samples accepted by CCAT would be evaluated using FLSS without sampling in latent space, while samples rejected by CCAT would be evaluated using FLSS with a threshold corresponding to the $10\%$ rejection criteria. Therefore, samples accepted by CCAT cannot be rejected by FLSS, ensuring that clean and correctly classified samples have a rejection threshold of $1\%$ similar to CCAT. Adversarial examples which are rejected by CCAT would have a higher accuracy, since they are predicted or rejected using FLSS which is adversarially trained. Images corrupted with low magnitude random noise, which are rejected by CCAT would be accurately predicted by FLSS, thereby improving the sensitivity of the overall system against random noise.  

The plots in Fig.\ref{fig:ccat} of the Supplementary show results against the adaptive attack considered by Stutz \etal \cite{stutz2019confidence}, which is a variant of maximum-margin loss coupled with momentum and backtracking. Against CCAT, this attack causes a high FW (accepted and incorrectly predicted images) in the range of $\delta\in[2/255,5/255]$, despite having a very high Rejection rate (R). In fact, rejection causes Robust Accuracy in CCAT to reduce, indicating that misclassified samples are not being rejected. Such low magnitude adversarial examples can be reliably predicted using FLSS, since it is adversarially trained. This results in a significant reduction in Rejection rate and boost in $Acc_{adv,10\%}$ when CCAT is combined with FLSS.

The results of the combined method (CCAT + FLSS) are presented for CIFAR-10 dataset in Table-\ref{table:main_all}. The clean accuracy corresponds to the No-Sampling case, while the adversarial accuracy is similar to FLSS ($SD=2$). The pre-trained model available on the CCAT GitHub repository has been used for CIFAR-10. We do not report CCAT results on CIFAR-$100$ since the pre-trained model is not available. Therefore, the method that combines FLSS with CCAT achieves the merits of both. We achieve the results similar to FLSS at a threshold corresponding to $1\%$ rejection of clean samples which are correctly predicted.

\subsection{Evaluation against EOT attack}
Since the proposed method utilizes randomization during inference, we evaluate its performance against Expectation over Transformation (EOT) attacks \cite{athalye2018obfuscated} (Table-\ref{table:eot} in the Supplementary). We forward propagate each input image $k$ times and use the average gradient direction to find the adversary. This is needed only for defenses which include randomization, and hence is not required for the baselines. We find that EOT attack is not stronger than the standard PGD attack, indicating that the gradients in the standard attacks are reliable to produce sufficiently strong attacks.  

\subsection{Checks to ensure absence of Gradient Masking}

We evaluate the model against stronger multi-step attacks and attacks with multiple random restarts and observe that the model trained using the proposed approach is stable to attacks with $1000$ steps and $10$ random restarts (Tables-\ref{table:cifar_restarts} and \ref{table:cifar100_restarts} in the Supplementary). Further, as suggested by Athalye \etal \cite{athalye2018obfuscated}, we perform the standard sanity checks and find that the proposed approach does not show gradient masking. 
We show evaluations of our model against the gradient-free attack, Square, in Tables-\ref{table:cifar_bb_all} and \ref{table:cifar100_bb_all} of the Supplementary. We note that gradient based attacks are stronger, thereby confirming the absence of gradient masking. We also evaluate our model against transfer-based black-box attacks (Tables-\ref{table:cifar_bb_all} and \ref{table:cifar100_bb_all} in the Supplementary) and note that they are significantly weaker than the other white-box attacks. 

\vspace{-0.05cm}
\section{Conclusions}
In this work, we seek to combine the desirable characteristics of adversarial training methods with detection techniques, while overcoming the deficiencies of both approaches. We propose adversarial training with Feature Level Stochastic Smoothing, wherein we utilise random smoothing of latent space features to obtain effectively smoother decision boundaries. Further, by enforcing the latent space to follow a fixed probability distribution during the training regime, we generate multiple predictions during inference, and subsequently assign a confidence score to each input sample based on consistency of predictions. By rejecting low confidence adversaries, we obtain a significant boost in performance on accepted samples over a wide range of attacks. Further, we demonstrate that by combining a popular adversarial detection method CCAT with the proposed approach, we achieve the merits of both methods. 

\section{Acknowledgements}
This work was supported by Uchhatar Avishkar Yojana (UAY) project (IISC\_10), MHRD, Govt. of India.

{
\bibliographystyle{ieee_fullname}
\bibliography{references}
}

\newpage
\appendix

\section{Training Details}

\begin{figure*}
\centering
        \includegraphics[width=\linewidth]{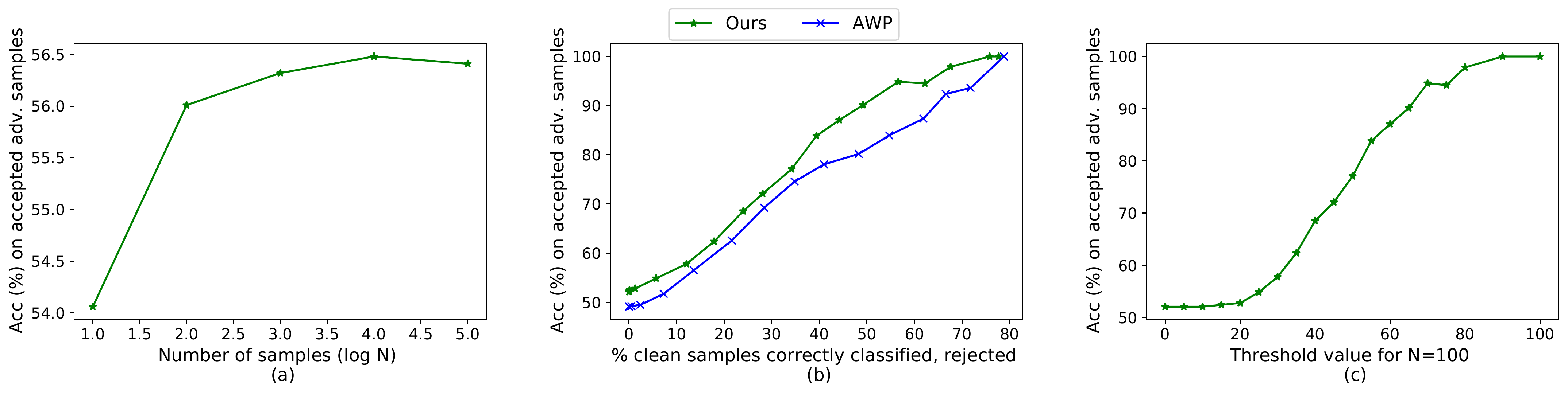}
        \caption{Plots on CIFAR-$10$ dataset (a) $Acc_{adv,10\%}$ is reported across variation in the number of times the latent vector is sampled during test time ($N$). The accuracy becomes stable with increase in $N$. $N$ is varied from $10^1$ to $10^5$. (b) Plot of $Acc_{adv,10\%}$ against variation in the percentage of clean samples that are correctly classified and rejected. Proposed method achieves improved accuracy compared to the strongest baseline, AWP \cite{wu2020adversarial}. (c)  $Acc_{adv,10\%}$ across variation in threshold value used for rejection. Threshold value of $32$ is selected to limit the percentage rejection of samples which are clean and correctly classified to $10\%$.}
        
        \label{fig:fpr}
\end{figure*}

We present details on the proposed training regime in this section. The proposed algorithm is presented in Algorithm-\ref{alg:train_algo} of the main paper.

We use coarse hyperparameter tuning (in exponents of $10$) to fix the coefficients weighting each loss term in Eq.\ref{eq:step1} and Eq.\ref{eq:step2} of the main paper. We set the weights of the Cross-Entropy loss ($\ell_{CE}$) and the KL divergence between latent space encodings of clean and adversarial samples ($KL_2$) to $1$. We assign weight of $1$ to the KL divergence between the softmax predictions of a clean image with and without sampling ($KL_4$) in L13 of Alg.1, so as to minimize the rejection of benign samples. Further, we set the weight of the KL divergence term that enforces encoder representations to obey the standard Gaussian distribution ($KL_1$) to $0.01$. Lastly, we utilise a weight of $0.1$ for the loss between the softmax predictions of adversarial images with and without latent space sampling ($KL_3$). We use the same set of hyperparameters across all datasets. 

We train the network for $120$ epochs, using a cyclic learning rate schedule. The maximum learning rate is set to $0.1$. We use Stochastic Gradient Descent (SGD) optimizer without momentum, and use a weight decay of $5$e-$4$.

We additionally perform early stopping using 7-step PGD adversarial samples on the designated validation set of each dataset, as explained in Section-\ref{datasets_and_arch}. We select the model which achieves the highest accuracy on adversarial samples in the \textit{No Sampling} case ($Acc_{adv,NS}$). We use early stopping for identifying the best models while training the baseline models as well.

\section{Rejection scheme}
\label{sec:rejection}
In this section, we discuss details on the rejection scheme described in Section-\ref{sub-sec:rejection} of the main paper. During test time, for each input image, the classifier predicts $N$ outputs after sampling $N$ times from the Gaussian distribution at the output of the encoder. We define the output of the \textit{Smoothed Classifier} to be the most frequently predicted class among the $N$ samples. Further, the sample is rejected if the frequency of the predicted class is below a predefined threshold $f$. We select the threshold for rejection such that not more than $10\%$ of the samples are correctly classified and rejected \cite{stutz2019confidence}. Based on this criteria, the threshold for the proposed method is set to $32$ for CIFAR-$10$ and $23$ for CIFAR-$100$. Therefore, for CIFAR-$10$ dataset, an input sample is accepted only if the classifier predicts the same output class at least $33$ times out of the $100$ outputs. 

Output of the ideal \textit{Smoothed Classifier} $SC$, is an expectation of the classifier outputs $C(x,\epsilon)$ over the random variable $\epsilon$, which is sampled from the Standard normal distribution. 
\begin{equation}
    SC(x) = \mathbb{E}_{\epsilon\sim\mathcal{N}(0,I)}~C(x,\epsilon)
\end{equation}

The above output $SC(x)$ is a deterministic value for every image $x$. However, in the absence of an analytical expression for this expectation, we consider sample statistics over all $N$ outputs obtained after sampling. This sample mean is not a deterministic value, and can vary during test time. A simple trick to make the outputs during test time deterministic is to pre-sample $N$ vectors from $\mathcal{N}(0,I)$, and use the same set of vectors during test time. Another option is to increase the value of $N$ during test time, which makes the predictions more stable to repeated evaluations. The plot of $Acc_{adv,10\%}$ vs. $N$ is shown in Fig.\ref{fig:fpr}(a). It can be seen that as the value of $N$ increases, the output becomes more stable. For all experiments in this paper, we consider $N=100$.

The metrics reported for any finite value of $N$ hold true with a fixed probability. In order to find this probability, we use the method of hypothesis testing as described in the work by Cohen \etal \cite{cohen2019certified,hung2019rank}. For each test sample, we find the $p$-value of the two-sided hypothesis test, $n_A\sim Binomial(n_A,n_B,p)$, where $n_A$ and $n_B$ represent the number of times the top-$2$ classes are predicted. We determine the value $\alpha$, which serves as an upper bound to this $p$-value for all test samples. Using this method, we find that for $N=100$, the probability of predicting an incorrect class for a sample which is claimed to be correct is $0.0045$. This value corresponds to the criterion of rejecting at most $10\%$ samples which are clean and correctly classified. We set the value of $N$ to be $100$ since the probability of incorrect prediction is sufficiently low, however this can be further reduced by increasing the value of $N$. 

As shown in Fig.\ref{fig:fpr}(c), the proposed \textit{Smoothed Classifier} can achieve a wide range of accuracies based on the threshold selected, at the cost of an increase in the fraction of samples which are clean, correctly classified and rejected. The plot in Fig.\ref{fig:fpr}(b) shows that for any given fraction of clean samples which are correctly predicted and rejected, the proposed method achieves higher adversarial accuracy on the accepted samples ($Acc_{adv,10\%}$), when compared to the AWP \cite{wu2020adversarial} baseline. 

\input{tables/cifar_1.tex}

\section{Details on Datasets and Model Architecture}
\label{datasets_and_arch}
To evaluate the proposed approach, we use the benchmark datasets, CIFAR-$10$ and CIFAR-$100$. CIFAR-$10$ \cite{krizhevsky2009learning} is a ten class dataset consisting of RGB images of dimension $32$ $\times$ $32$, and is commonly used to benchmark results on adversarial robustness of deep networks. The original training set consists of $50,000$ images, which we split into $49,000$ images to comprise the training set, and a hold-out set of $1,000$ images to serve as the validation set. Using this validation set, we perform early-stopping to identify the best model parameters for the proposed approach as well the baseline defense methods. 
Further, to demonstrate the scalability of the proposed approach to datasets with higher number of classes, we present results on CIFAR-$100$, which is a $100$-class dataset with RGB images of dimension $32$ $\times$ $32$. We use a class-balanced validation set of size $2500$ for CIFAR-$100$.

We use ResNet-$18$ architecture to report the performance of the proposed approach as well as other existing defenses such as AWP \cite{wu2020adversarial}, TRADES \cite{zhang2019theoretically} and PGD \cite{madry-iclr-2018}. However, since we use the pre-trained model for CCAT \cite{stutz2019confidence}, the architecture is ResNet-$20$, as used by the authors. The CNN backbone of ResNet-$18$ forms the encoder network of the proposed approach. We require an additional layer with $1\times1$ convolutions to compute the $512$ dimensional mean and variance vectors in the proposed approach when compared to the baselines. Therefore, the architecture of the proposed approach uses ~$5\%$ additional parameters when compared to the baselines. The MLP in Fig.\ref{fig:network-architecture} of the main paper consists of 3 layers. However, to ensure that the architecture of the proposed method is close to the ResNet-$18$ architecture used for baselines, we use 1 layer in the MLP head. We obtain slight gains (close to $0.5\%$) in robust accuracy by including an additional layer in the MLP head.

\section{Experimental results}

We present more details on evaluation of the proposed method, and comparison of results against baselines in this section. We use Nvidia DGX workstation with V100 GPUs for our training and evaluation. 
\input{tables/ablations}

\subsection{Ablation Experiments}

We present results on various ablation experiments in Table-\ref{table:ablations}, to highlight the importance of different components of the proposed algorithm. In S1, we skip training on the second step (Eq.\ref{eq:step2} of the main paper), which is important for improving the accuracy on clean samples. While this results in a $2\%$ boost in adversarial accuracy after rejection ($Acc_{adv,10\%}$), the clean accuracy drops with respect to the proposed approach (P). In S2, we train on a combination of the losses in Eq.\ref{eq:step1} and \ref{eq:step2} of the main paper. While this saves the computation time for one additional back propagation, the adversarial accuracy drops. It is possible that careful tuning of hyperparameters can retrieve the best accuracy using the combined loss. 

We further set each of the KL divergence terms to 0, one at a time, in A1-A4. While all these experiments result in sub-optimal results, setting KL2 to $0$ causes a very significant drop in accuracy on adversarial samples, since it directly relates to the robustness objective. We also try to replace KL2 with $0.1 \cdot KL\big(E(\widetilde{x})\vert\vert \mathcal{N}(0,I))$, which also leads to sub-optimal robustness. As seen in A6, increasing the coefficient of KL1 results in a higher clean accuracy, at the cost of a drop in accuracy of adversarial samples. Increasing KL3 from $0.1$ to $1$ (A6) also leads to a drop in robustness.

\begin{figure*}
\centering
        \includegraphics[width=\linewidth]{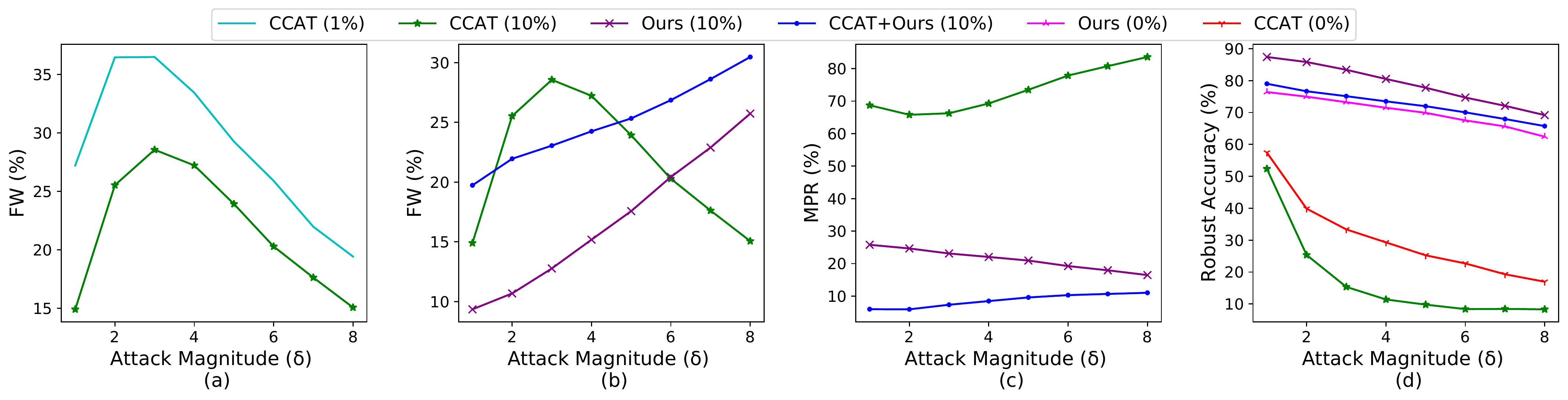}
        \caption{Performance against an enhanced Maximum-Margin attack \cite{stutz2019confidence} on CIFAR-$10$ dataset against CCAT and FLSS (Ours) models (a) Plot of FW on CCAT against attack magnitude ($\delta$) at different FPR values (b) Comparison of FW between CCAT, FLSS (Ours) and CCAT + FLSS (Ours) (c) Maximum Percentage Rejection ($\%$) for the same models (d) Comparison of Robust accuracy with and without rejection}
        \label{fig:ccat}
\end{figure*}

\input{tables/CCAT_randnoise}

\subsection{Combining FLSS with CCAT}

We present the performance of the state-of-the-art Adversarial Detection method CCAT \cite{stutz2019confidence} against random perturbations sampled from a Bernoulli distribution of varying magnitudes (denoted by $\delta$) in Table-\ref{table:ccat_noise}. Since the goal of Adversarial Detection methods is to be able to detect images which do not belong to the original distribution, they are seen to be overly sensitive to even perturbations of small magnitude. We note that even at $\delta=1/255$, the network rejects $78.6\%$ of the images at a rejection threshold of $1\%$. The accuracy of a normally trained ResNet-$18$ network is indeed very high ($92.4\%$) against images corrupted with such low magnitude ($\delta=1/255$) random noise sampled from a Bernoulli distribution. In order to address this sensitivity, we propose to combine FLSS with CCAT such that every image that is rejected by CCAT is re-evaluated for acceptance by FLSS. This reduces the rejection to $25.4\%$, while maintaining a high accuracy of $89.49\%$ after rejection. For the combined model we consider a threshold corresponding to $1\%$ rejection of correctly classified clean samples for CCAT, and $10\%$ for FLSS, which applies to the samples rejected by CCAT. Therefore the effective criteria for the combined model is very close to $1\%$. 

We further evaluate the performance of CCAT, FLSS and a combination of both approaches against an enhanced Maximum Margin attack proposed by Stutz \etal \cite{stutz2019confidence} in Fig.\ref{fig:ccat}. As noted by the authors, this attack causes a higher rate of misclassification on accepted samples (FW) for CCAT when compared to other attacks. As seen in Fig.\ref{fig:ccat}(a), with a threshold corresponding to $1\%$ rejection of clean samples that are correctly classified, FW is $36.5\%$. This reduces to $28.56\%$ when the rejection threshold is increased to $10\%$. For the same threshold of $10\%$, the proposed method has a worst case FW of $25.73\%$ at $\delta=8/255$ as shown in Fig.1(b). This is achieved at a significantly lower Maximum Percentage Rejection (MPR) as shown in Fig.\ref{fig:ccat}(c). While FW increases to $30.47\%$ when combined with FLSS, this is again achieved at a very low MPR. We present the Robust Accuracy after rejection ($Acc_{adv,10\%}$) in Fig.\ref{fig:ccat}(d). Since the CCAT model is not adversarially trained, the accuracy is very low. However, when combined with FLSS, there is a significant boost in accuracy.

\subsection{Evaluation against Black-Box attacks}
We evaluate the proposed method against transfer based black-box attacks on CIFAR-10 and CIFAR-100 in Tables-\ref{table:cifar_bb_all} and \ref{table:cifar100_bb_all} respectively. Here, we consider FGSM and PGD $7$-step transfer-based attacks, as well as the query-based Square attack \cite{andriushchenko2019square}. For the transfer attacks, the source model considered is a normally trained model of the same architecture as the target network. We note that FGSM black-box attack is stronger than PGD $7$-step attack, while the Square attack is the strongest as expected, since it performs zeroth-order optimization on the model. We also observe that the proposed method performs significantly better than AWP on the strongest attack (Square), for both $0\%$ as well as $10\%$ rejection rates. We note that black-box attacks are significantly weaker than the white-box attacks reported in the main paper. This shows the absence of gradient masking in the proposed method.

\input{tables/eot}
\input{tables/cifar_bb_fgsm}
\input{tables/cifar100_bb_fgsm}

\subsection{Evaluation using Random Restarts}
\label{rr}
For reliable evaluation of the proposed defense, we present results against multiple random restarts and multiple steps of the PGD attack on CIFAR-10 and CIFAR-100 datasets in Tables-\ref{table:cifar_restarts} and \ref{table:cifar100_restarts}. The purpose of multiple restarts is to find the worst adversary in the $\delta$-ball around each image. It is to be noted that a naive implementation of a series of attacks on a single image could lead to an inadvertent drop in accuracy due to the stochasticity of predictions. In fact, even if a single clean image is repeatedly evaluated ($n$ times) on the proposed classifier, it is likely to be rejected at least once as $n \to \infty$. We capture the same by reporting the probability of correct predictions using hypothesis testing in Section-\ref{sec:rejection}. However, since the goal in this section is to merely find an adversary in the $\delta$-ball of each image, we consider a fixed set of noise vectors sampled from $\mathcal{N}(0,I)$ for each attack in Tables-\ref{table:cifar_restarts} and \ref{table:cifar100_restarts}. The same is considered for all other evaluations in the paper which use a series of attacks on a single image.

We observe from the first two partitions of Tables-\ref{table:cifar_restarts} and \ref{table:cifar100_restarts} that there is only a marginal drop in robust accuracy between the PGD-100 and PGD-$1000$ step attacks. This indicates that the robust accuracy saturates and does not deteriorate further as the number of steps used in the PGD attack is increased. Further, from the first and third  partitions, we observe that the robust accuracy is preserved even with multiple random restarts of the PGD attack, thereby indicating the absence of the gradient masking effect.

\input{tables/cifar_restarts}
\input{tables/cifar100_restarts}
\input{tables/adaptive_cifar10}

\begin{figure*}
\centering
        \includegraphics[width=\linewidth]{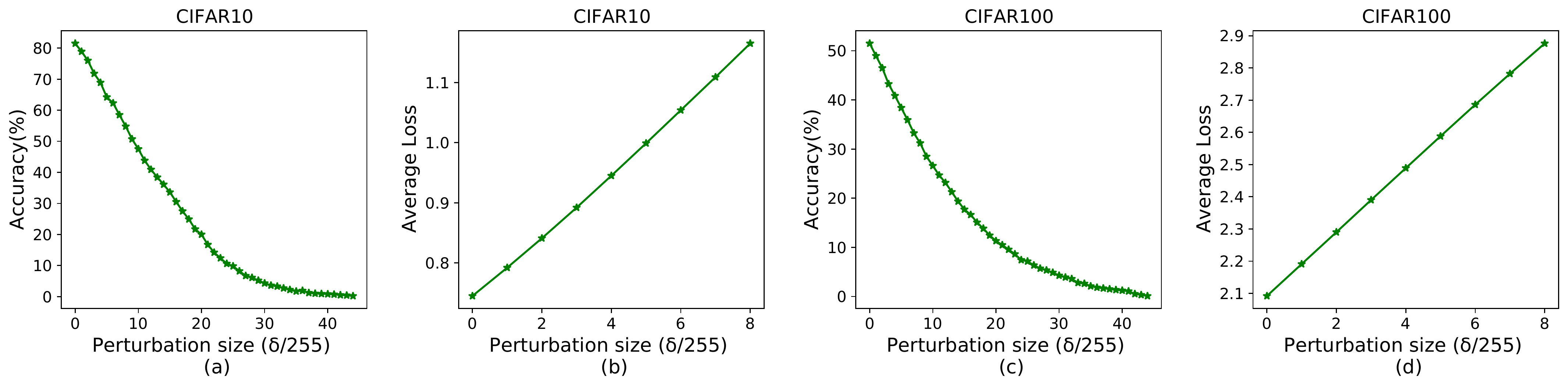}
        \caption{Accuracy and loss on adversaries in a no-sample case (a) Accuracy ($Acc_{adv,NS}$) against PGD-$7$ step attack on CIFAR-$10$  dataset. (b) Loss on FGSM adversaries for CIFAR-$10$ dataset. (c) Accuracy ($Acc_{adv,NS}$) against PGD-$7$ step attack on CIFAR-$100$  dataset. (d) Loss on FGSM adversaries for CIFAR-$100$ dataset. We note that $Acc_{adv,NS}$ goes to $0$ for higher perturbation magnitudes, and loss on FGSM samples increases monotonically with $\delta$, indicating the absence of gradient masking.}
        \label{fig:acc_loss}
\end{figure*}

\subsection{Evaluation against Adaptive attacks}

We evaluate our proposed model against various adaptive adversarial attacks, which are constructed specifically for the defense at hand, as recommended by Tramer \etal \cite{tramer2020adaptive}. The results for CIFAR-$10$ are reported in Table-\ref{table:cifar_adaptive}. We use a $1000$-sample balanced subset for reporting our results. 

We broadly consider two kinds of attacks. The first set of attacks, A$1$ to A$6$ are implemented with the objective of fooling the model, whereas the second set of attacks, RA$1$ to RA$4$ are crafted to encourage the model to reject the image. For the proposed model, we claim improved adversarial accuracy on accepted samples ($Acc_{adv,10\%}$), while maintaining a reasonable limit on the maximum percentage rejection (MPR). Therefore performance against both types of attacks needs to be considered. In most of the attacks (unless specified), the image is sampled once during the forward propagation for attack generation. 

Since the proposed training method relies on the properties of the feature space of an image, we consider a wide range of feature level attacks (FA) at the output of the encoder, to generate an adaptive adversary. In A$1$ and A$2$, we craft an adversary that closely resembles a different image from a target class in the feature space. A$1$ crafts an adversary to minimize the KL divergence between the encoder outputs of the two images, whereas A$2$ crafts an adversary to minimize the MSE between the predicted mean vectors. For A$1$ and A$2$, we consider a random image from each of the $10$ classes, and report the worst case accuracy across all attacks. We find that A$1$ is the strongest among all the adaptive attacks considered, yielding an accuracy of $69.71\%$ on accepted adversarial samples for CIFAR-$10$ dataset. 
A$3$ and A$4$ are untargeted versions of the attacks A$1$ and A$2$ respectively. These attacks generate an adversary to maximize the KL divergence and MSE with respect to the corresponding clean images in the latent space. 

A$5$ optimizes an objective which is similar to A$2$, however, it also attempts to find an adversary with minimum variance at the output of the encoder. Having a low variance enforces the outputs of all random samples at test time to be similar, thereby encouraging the adversary to be accepted even if it is incorrectly classified. In A$6$, each of the $100$ outputs of the test image are encouraged to be predicted as a fixed random target class, by minimizing cross-entropy loss of each of the softmax vectors with respect to this target. The overall loss which is optimized is the sum of all $100$ cross-entropy losses. 

The next set of adaptive attacks consider the objective of increasing MPR (Maximum Percentage Rejection). RA$1$ generates an adversary that maximizes entropy of the output probability vector. This would possibly diversify the prediction of the image when sampled multiple times, thereby resulting in rejection of the image. In RA$2$, we consider the same objective of maximizing entropy, but additionally enforce that the image is correctly predicted, by minimizing cross-entropy of the image with respect to the true class. We find that RA$1$ is the strongest attack, leading to $50.8\%$ rejection. In RA$3$, each of the $100$ sampled outputs of the test image are encouraged to be predicted as a different random class. This is done by minimizing cross-entropy loss of each of the softmax vectors with respect to a different random target, and finding the gradient of sum of all losses. This attack is marginally weaker than RA$1$ and RA$2$, possibly because of inconsistency in the loss, resulting in a weak gradient direction. In RA$4$, an adversary is generated by maximizing variance at the output of the encoder. Higher variance can possibly lead to inconsistent predictions, thereby leading to higher rejection rate. However, we find this to be weaker than the other attacks. 

Overall, we find that the ensemble of $6$ attacks considered in Table-\ref{table:main_all} of the main paper are significantly stronger than attacks which can possibly exploit the specific nature of the defense. The attack RA-$1$ however is able to increase the Rejection rate higher than the ensemble of attacks considered for the main evaluations in the paper. We therefore consider the RA-$1$ attack on the CIFAR-$100$ dataset as well. This increases the MPR on CIFAR-$100$ from $65.61\%$ to $72.28\%$, while maintaining a significantly high accuracy on adversarial samples before and after rejection.

\subsection{Sanity Checks to ensure absence of Gradient Masking}

The plots in Fig.\ref{fig:acc_loss} show that for both CIFAR-$10$ and CIFAR-$100$, accuracy against PGD $7$-step attack goes to $0$ as the magnitude of perturbation ($\delta$) increases. Also, loss on FGSM adversaries for small perturbation magnitudes increases monotonically. Both of these trends indicate the absence of gradient masking in the proposed method \cite{athalye2018obfuscated}. We consider the \textit{No Sampling} case here, since this is primarily a check to verify the efficacy of the defense, and hence the rejection scheme need not be considered.

\end{document}

%% file: Training_Algo.tex
\begin{algorithm}[tb]
   \caption{Adversarial Training using Feature Level Stochastic Smoothing Classifier}
   \label{alg:train_algo}
   
\begin{algorithmic}[1]
   \STATE {\bfseries Input:} Classifier Network $C_{\theta}$ (with parameters $\theta$, Encoder $E= \{ \mu_{\theta_{E}}, \Sigma_{\theta_E}\}$, MLP $M$), Training Data $\{x_i,y_i \}_{i=1}^{K}$, Epochs $T$, Learning Rate $\eta$, Adversarial Perturbation function $A(C_\theta(x,\epsilon),y)$
\FOR{$epoch=1$ {\bfseries to} $T$}
    \FOR{$i=1$ {\bfseries to} $K$}
        \STATE Sample $\epsilon \sim \mathcal{N}(0,I)$
        \STATE $\widetilde{x}_i = A(C_\theta(x_i,\epsilon),y_i)$
        \STATE $KL_1 = KL\big( \mathcal{N}(E(x_i)) \vert\vert \mathcal{N}(0,I)  \big)$
        \STATE $KL_2 = KL\big( \mathcal{N}(E(\widetilde{x}_i)) \vert\vert \mathcal{N}(E(x_i))  \big)$
        \STATE $KL_3 = KL \big( C_\theta(\widetilde{x}_i,\epsilon) \vert\vert C_\theta(\widetilde{x}_i,0) \big)$
        \STATE $L = \ell_{CE}(C_{\theta} (\widetilde{x}_i,\epsilon) , y_i ) + KL_1 + KL_2 + KL_3$
        
        \STATE $\theta = \theta -  \eta \cdot \nabla_{\theta} L   $
        \\ 
        \STATE Sample $\epsilon' \sim \mathcal{N}(0,I)$
        \STATE $KL_1 = KL\big( \mathcal{N}(E(x_i)) \vert\vert \mathcal{N}(0,I)  \big)$
        \STATE $KL_4 =  KL(C_{\theta} (x_i,\epsilon') \vert\vert C_{\theta} (x_i,0))$
        \STATE $L = \ell_{CE}(C_{\theta} (x_i,\epsilon') , y_i ) + KL_1 + KL_4 $
        \STATE $\theta = \theta -  \eta \cdot \nabla_{\theta} L   $

    \ENDFOR

\ENDFOR
\end{algorithmic}
\end{algorithm}

%% file: tables/main_all.tex
\begin{table*}[]
\caption{\textbf{White-Box Evaluation}: Performance (\%) of models under an ensemble of 6 attacks : PGD, APGD-CE, APGD-DLR \cite{croce2020reliable}, PGD-CW \cite{carlini2017towards}, GAMA-PGD and GAMA-MT \cite{sriramanan2020gama}. FC denotes the $\%$ of samples which are correctly classified and accepted for all attacks. FW denotes the $\%$ of samples which are accepted and incorrectly classified by at least one attack. MPR denotes the max $\%$ rejected samples.}

\label{table:main_all}
\vspace{0.05cm}
\setlength\tabcolsep{3pt}
\resizebox{1.0\linewidth}{!}{
\begin{tabular}{l|c|ccc|ccc|ccc}
\toprule
        Method          & Thresholding & \textbf{$Acc_{nat,NS}\uparrow$} & \textbf{$Acc_{nat,0\%}\uparrow$} & \textbf{$Acc_{nat,10\%}\uparrow$} & \textbf{$Acc_{adv,NS}\uparrow$} & \textbf{$Acc_{adv,0\%}\uparrow$} & \textbf{$Acc_{adv,10\%}\uparrow$} & FC $\uparrow$         & FW  $\downarrow$       & MPR         \\
   \midrule    
\multicolumn{11}{c}{\textbf{CIFAR-10}}                   \\                                 \midrule                                                                                                                                                 
PGD-AT \cite{madry-iclr-2018,rice2020overfitting,pang2020bag}                & Confidence            & 83.80                    & 83.80                     & 91.93                      & 49.07                    & 49.07                     & 51.15                      & 43.99                & 42.00                & 44.19                \\
TRADES \cite{zhang2019theoretically}            & Confidence            & 81.77                    & 81.77                     & 90.26                      & 49.43                    & 49.43                     & 51.85                      & \textbf{44.13}       & 40.97                & 44.34                \\
AWP \cite{wu2020adversarial}               & Confidence            & 80.58                    & 80.58                     & 89.14                      & 49.80                    & 49.80                     & 53.01                      & 44.06                & 39.05                & 44.01                \\
CCAT \cite{stutz2019confidence}               & Confidence            & \textbf{89.92}           & \textbf{89.92}            & \textbf{97.52}             & 0.00                     & 0.00                      & 0.00                       & 0.00                 & \textbf{8.52}        & 100.00               \\
FLSS (\textbf{Ours}) (SD=1)      & Confidence            & 80.51                    & 80.51                     & 89.10                      & \textbf{50.64}           & 50.64                     & 54.06                      & 42.28                & 35.84                & 42.18       \\
FLSS (\textbf{Ours}) (SD=2) & Maj. Vote             & 80.51                    & 77.68                     & 89.63                      & \textbf{50.64}           & \textbf{51.00}            & \textbf{56.16}             & 43.16                & 33.69                & 47.42                \\
CCAT \cite{stutz2019confidence} + FLSS (\textbf{Ours}) & Conf + Maj. Vote      & 80.51                    & \textbf{-}                & 89.10                      & \textbf{50.64}           & \textbf{-}                & \textbf{56.16}             & 43.16                & 33.69                & 47.42                \\
\midrule
\multicolumn{11}{c}{\textbf{CIFAR-100}}     \\                                          \midrule                                                                                                                                                                                        
PGD-AT \cite{madry-iclr-2018,rice2020overfitting,pang2020bag}                & Confidence            & 56.13                    & 56.13                     & 74.30                      & 25.40                    & 25.40                     & 26.06                      & 23.30                & 66.08                & 59.08                \\
TRADES \cite{zhang2019theoretically}           & Confidence            & 57.84                    & 57.84                     & 74.09                      & 24.33                    & 24.33                     & 23.84                      & 22.70                & 72.49                & 57.12                \\
AWP \cite{wu2020adversarial}               & Confidence            & \textbf{58.21}           & \textbf{58.21}            & 74.31                      & 25.16                    & 25.16                     & 25.07                      & 23.43       & 70.01                & 55.77                \\
FLSS (\textbf{Ours}) (SD=1)       & Confidence            & 51.86                    & 51.86                     & 70.86                      & \textbf{25.57}                    & \textbf{25.57}                     & 28.79                      & \textbf{23.51}          & 58.14         & 56.48 \\
FLSS (\textbf{Ours}) (SD=2) & Maj. Vote             & 51.86                    & 47.50                     & \textbf{74.35}             & \textbf{25.57}           & \textbf{25.57}            & \textbf{29.95}             & 22.01                & \textbf{51.46}       & 65.61                \\
\bottomrule
\end{tabular}
}
\vspace{-0.3cm}
\end{table*}

%% file: tables/cifar_1.tex
\begin{table}[]
\caption{\textbf{CIFAR-$10$}: Performance (\%) of models under an ensemble of 5 attacks : PGD, APGD-CE, APGD-DLR, FAB, SQUARE. FC denotes the percentage of samples which are always correctly classified and accepted for all attacks. FW denotes the percentage of samples which are accepted and incorrectly classified by at least one of the attacks. MPR denotes the max $\%$ rejected samples.}
\label{table:cifar_1}
\vspace{-0.0cm}
\setlength\tabcolsep{3pt}
\resizebox{1.0\linewidth}{!}{
\begin{tabular}{l|ccccc}
\toprule
\multicolumn{1}{l|}{}                & \small{$Acc_{nat, NS}\uparrow$} & \small{$Acc_{adv, 10\%}\uparrow$}  & FC$\uparrow$ & FW$\downarrow$ & MPR \\
\midrule
RS standard training \cite{cohen2019certified}      & \textbf{86.32} & 16.17   & 14.47 & 75.00    & 58.16 \\
PNI-W (Noise)  \cite{he2019parametric}               & 85.48 & 34.69  & 25.31 & 47.67 & 64.79 \\
Trades+RS (Noise) \cite{blum2020random} & 75.14 & 49.28  & 41.48 & 42.69 & 44.52 \\

PGD (Noise) \cite{madry-iclr-2018}                     & 84.37      & 51.63                            & 41.88        & 39.23        & 45.32      \\
TRADES (Noise) \cite{zhang2019theoretically}                  & 80.15      & 50.83                                          & 42.24        & 40.85        & 40.92                   \\
PGD (Conf)  \cite{madry-iclr-2018}     & 83.80 & 59.40 & 44.03 & 30.09 & 41.20 \\
Trades (Conf) \cite{zhang2019theoretically}    & 81.77 & 62.10 & 44.15 & 26.94 & 41.19 \\
AWP (Conf) \cite{wu2020adversarial} & 80.58 & 63.38 & 44.06 & 25.45 & 40.55 \\
CCAT \cite{stutz2019confidence} & 89.92 & 0.00 & 0.00 & \textbf{1.85} & 100.00 \\
FLSS (\textbf{Ours})             & 80.22 & \textbf{72.40} & \textbf{51.10} & 19.46 & \textbf{37.69} \\
\bottomrule
\end{tabular}
}
\end{table}

%% file: tables/ablations.tex
\begin{table*}[]
\caption{\textbf{Ablations on CIFAR-10}: Performance (\%) of models under an ensemble of 6 attacks : PGD, APGD-CE, APGD-DLR \cite{croce2020reliable}, PGD-CW \cite{carlini2017towards}, GAMA-PGD and GAMA-MT \cite{sriramanan2020gama}. FC denotes the percentage of samples which are always correctly classified and accepted for all attacks. FW denotes the percentage of samples which are accepted and incorrectly classified by at least one of the attacks. MPR denotes the max $\%$ rejected samples.}
\label{table:ablations}
\vspace{-0.0cm}
\setlength\tabcolsep{3pt}
\resizebox{1.0\linewidth}{!}{
\begin{tabular}{l|l|ccc|ccc|ccc}
\toprule
                  & Description & \textbf{$Acc_{nat,NS}\uparrow$} & \textbf{$Acc_{nat,0\%}\uparrow$} & \textbf{$Acc_{nat,10\%}\uparrow$} & \textbf{$Acc_{adv,NS}\uparrow$} & \textbf{$Acc_{adv,0\%}\uparrow$} & \textbf{$Acc_{adv,10\%}\uparrow$} & FC $\uparrow$         & FW  $\downarrow$       & MPR         \\
   \midrule    
P  & Our Proposed Approach                                           & 80.51 & 77.68 & 89.63 & 50.64 & 51.00 & 56.16 & 43.16 & 33.69 & 47.42 \\

S1 & 1-step training (Skip Eq.2 of main paper)           & 72.16 & 65.81 & 86.31 & 53.16 & 48.10 & 58.19 & 37.47 & 26.92 & 53.30 \\
S2 & 1-step training (Combine Eq.1, 2 of main paper) & 80.76 & 78.31 & 89.25 & 50.51 & 49.45 & 54.22 & 42.35 & 35.75 & 46.97 \\
\midrule
A1 & Set coefficient of KL1 to 0                         & 80.08 & 77.42 & 89.41 & 53.49 & 49.17 & 54.84 & 41.66 & 34.30 & 47.86 \\
A2 & Set coefficient of KL2 to 0                                     & 84.31 & 84.07 & 91.88 & 40.60 & 41.31 & 41.92 & 36.81 & 50.98 & 30.05 \\
A3 & Set coefficient of KL3 to 0                                     & 79.12 & 77.71 & 87.95 & 53.41 & 48.91 & 52.42 & 42.64 & 38.70 & 46.08 \\
A4 & Set coefficient of KL4 to 0                                     & 81.05 & 77.99 & 88.62 & 50.92 & 50.93 & 53.85 & 44.81 & 38.40 & 44.02 \\
A5 & Increase coefficient of KL1 from 0.01 to 0.1                    & 81.33 & 74.51 & 89.97 & 50.11 & 50.56 & 55.25 & 41.53 & 33.62 & 47.06 \\
A6 & Increase coefficient of KL3 from 0.1 to 1                       & 80.34 & 76.78 & 89.02 & 50.70 & 51.16 & 55.04 & 43.95 & 35.90 & 45.43 \\
A7 & Replace KL2 with $0.1 \cdot KL (adv || \mathcal{N}(0,1))$                       & 85.01 & 84.09 & 92.90 & 43.73 & 44.01 & 45.64 & 39.42 & 46.94 & 49.58    \\
\bottomrule
\end{tabular}
}
\end{table*}

%% file: tables/CCAT_randnoise.tex
\begin{table}[]
\caption{\textbf{Performance ($\%$) of CCAT \cite{stutz2019confidence}} against random Bernoulli noise perturbations of varying magnitude (denoted by $\delta$). The rejection is done such that not more than $1\%$ of the clean samples are correctly classified and rejected. FC denotes the percentage of samples which are always correctly classified and accepted for all attacks. FW denotes the percentage of samples which are accepted and incorrectly classified by at least one of the attacks. R denotes the $\%$ rejected samples. CCAT rejects a high fraction of samples even at low perturbation magnitudes.}
\label{table:ccat_noise}
\vspace{-0.0cm}
\setlength\tabcolsep{3pt}
\resizebox{1.0\linewidth}{!}{
\begin{tabular}{c|ccccc}
\toprule
        Noise ($\delta$) & ~~~\textbf{$Acc_{0\%}\uparrow$}~~~ & ~~~\textbf{$Acc_{1\%}\uparrow$}~~~  & ~~~FC$\uparrow$~~~         & ~~~FW  $\downarrow$~~~       & ~~~R~~~         \\
   \midrule    
$1/255$                & 75.82   & 97.99   & 20.97 & 0.43 & 78.60  \\
$2/255$                & 40.04   & 97.43   & 1.14  & 0.03 & 98.83  \\
$3/255$                & 26.03   & 90.90   & 0.10  & 0.01 & 99.89  \\
$4/255$                & 21.06   & 66.66   & 0.02  & 0.01 & 99.97  \\
$5/255$                & 17.75   & 0.00    & 0.00  & 0.01 & 99.99  \\
$6/255$                & 16.21   & -       & 0.00  & 0.00 & 100.00 \\
$7/255$                & 16.03   & -       & 0.00  & 0.00 & 100.00 \\
$8/255$                & 15.96   & -       & 0.00  & 0.00 & 100.00\\
\bottomrule
\end{tabular}
}
\end{table}

%% file: tables/eot.tex
\begin{table}[]
\caption{\textbf{EOT}: Accuracy (\%) of models against Expectation over Transformation (EOT) attack \cite{athalye2018obfuscated} on CIFAR-10 and CIFAR-100 datasets. The base attack considered is PGD-$100$. EOT-$k$ represents the use of $k$ computations to approximate the expected value of the gradient. We report results for $k=10$, $50$ and $100$. The accuracy of the proposed approach is stable to EOT attacks.}
\label{table:eot}
\vspace{0.1cm}
\setlength\tabcolsep{3pt}
\resizebox{1.0\linewidth}{!}{
\begin{tabular}{l|cc|cc}
\toprule
\multicolumn{1}{l}{} & \multicolumn{2}{|c}{CIFAR-10}                             & \multicolumn{2}{|c}{CIFAR-100}                            \\
\multicolumn{1}{l}{} & \multicolumn{1}{|l}{$Acc_{adv,0\%}$} & \multicolumn{1}{l}{$Acc_{adv,10\%}$} & \multicolumn{1}{|l}{$Acc_{adv,0\%}$} & \multicolumn{1}{l}{$Acc_{adv,10\%}$} \\
\midrule
Standard Attack      & 54.00                         & 65.65                       & 29.16                      & 47.83                       \\
EOT - 10             & 54.40                       & 66.61                       & 30.16                      & 53.96                       \\
EOT - 50             & 54.60                       & 66.82                       & 30.00                         & 53.95                       \\
EOT - 100            & 54.60                       & 67.12                       & 30.08                      & 54.01    \\
\bottomrule
\end{tabular}
}
\end{table}

%% file: tables/cifar_bb_fgsm.tex
\begin{table}[]
\caption{\textbf{CIFAR-$10$}: Accuracy (\%) of models against FGSM, PGD-7 and Square Black-Box attacks. Attack source for FGSM and PGD-7 is a normally trained model of the same architecture.}
\label{table:cifar_bb_all}
\vspace{0.1cm}
\setlength\tabcolsep{3pt}
\resizebox{1.0\linewidth}{!}{
\begin{tabular}{l|cc|cc}
\toprule
\multicolumn{1}{l|}{} & \textbf{$Acc_{nat,0\%}\uparrow$} & \textbf{$Acc_{nat,10\%}\uparrow$} & \textbf{$Acc_{adv,0\%}\uparrow$} & \textbf{$Acc_{adv,10\%}\uparrow$} \\
\midrule
AWP - FGSM    & 80.58 & 89.14 & 78.01 & 87.23 \\
Ours - FGSM   & 77.68 & 89.63 & 75.98 & 88.96 \\
AWP - PGD 7   & 80.58 & 89.14 & 78.86 & 88.02 \\
Ours - PGD 7  & 77.68 & 89.63 & 76.32 & 89.12 \\
AWP - Square  & 80.58 & 89.14 & 55.52 & 78.69 \\
Ours - Square & 77.68 & 89.63 & 70.40  & 85.16           \\
\bottomrule
\end{tabular}
}
\end{table}

%% file: tables/cifar100_bb_fgsm.tex
\begin{table}[]
\caption{\textbf{CIFAR-$100$}: Accuracy (\%) of models against FGSM, PGD-7 and Square Black-Box attacks. Attack source for FGSM and PGD-7 is a normally trained model of the same architecture.}
\label{table:cifar100_bb_all}
\vspace{0.1cm}
\setlength\tabcolsep{3pt}
\resizebox{1.0\linewidth}{!}{
\begin{tabular}{l|cc|cc}
\toprule
\multicolumn{1}{l|}{} & \textbf{$Acc_{nat,0\%}\uparrow$} & \textbf{$Acc_{nat,10\%}\uparrow$} & \textbf{$Acc_{adv,0\%}\uparrow$} & \textbf{$Acc_{adv,10\%}\uparrow$} \\
\midrule
AWP - FGSM    & 58.21 & 74.31 & 56.52 & 73.01 \\
Ours - FGSM   & 47.50 & 74.35 & 46.03 & 73.21 \\
AWP - PGD 7   & 58.21 & 74.31 & 56.78 & 73.56 \\
Ours - PGD 7  & 47.50 & 74.35 & 46.15 & 73.62 \\
AWP - Square  & 58.21 & 74.31 & 30.96 & 53.27 \\
Ours - Square & 47.50 & 74.35 & 40.35 & 70.75     \\
\bottomrule
\end{tabular}
}
\end{table}

%% file: tables/cifar_restarts.tex
\begin{table}[]
\caption{\textbf{CIFAR-$10$}: Accuracy (\%) of models under attacks with varying number of steps and restarts. Accuracy with no rejection and $10\%$ rejection is reported for each attack.}
\label{table:cifar_restarts}
\vspace{-0.0cm}
\setlength\tabcolsep{3pt}
\resizebox{1.0\linewidth}{!}{
\begin{tabular}{l|cc|cc|cc}
\toprule
                & $Acc_{0\%}\uparrow$                                & $Acc_{10\%}\uparrow$                                & $Acc_{0\%}\uparrow$                                 & $Acc_{10\%}\uparrow$                                 & $Acc_{0\%}\uparrow$                        & $Acc_{10\%}\uparrow$ \\
                \midrule
No. of steps    & 100                                     & 100                                      & 1000                                     & 1000                                      &     100                            &                      100            \\

No. of restarts & 1                                       & 1                                        & 1                                        & 1                                         & 10                              & 10                               \\
\midrule
\midrule
AWP \cite{wu2020adversarial} & 53.92                & 66.36                & 53.87                & 66.30                & 53.92                & 66.30                \\
FLSS (\textbf{Ours}) & 54.61                & 67.89                & 54.55                & 67.88                & 54.57                & 67.72                \\
\bottomrule
\end{tabular}}
\end{table}

%% file: tables/cifar100_restarts.tex
\begin{table}[]
\caption{\textbf{CIFAR-$100$}: Accuracy (\%) of models under attacks with varying number of steps and restarts. Accuracy with no rejection and $10\%$ rejection is reported for each attack.}
\label{table:cifar100_restarts}
\vspace{-0.0cm}
\setlength\tabcolsep{3pt}
\resizebox{1.0\linewidth}{!}{
\begin{tabular}{l|cc|cc|cc}
\toprule
                & $Acc_{0\%}\uparrow$                                & $Acc_{10\%}\uparrow$                                & $Acc_{0\%}\uparrow$                                 & $Acc_{10\%}\uparrow$                                 & $Acc_{0\%}\uparrow$                        & $Acc_{10\%}\uparrow$ \\
                \midrule
No. of steps    & 100                                     & 100                                      & 1000                                     & 1000                                      &     100                            &                      100            \\

No. of restarts & 1                                       & 1                                        & 1                                        & 1                                         & 10                              & 10                               \\
\midrule
\midrule
AWP \cite{wu2020adversarial} & 31.28                & 45.66                & 31.24                & 45.86                & 31.11                & 45.51               \\
FLSS (\textbf{Ours}) & 29.16                & 48.00                & 29.08                & 47.88                & 29.01                & 47.62                \\
\bottomrule
\end{tabular}}
\end{table}

%% file: tables/adaptive_cifar10.tex
\begin{table}[]
\caption{\textbf{Adaptive Attacks on CIFAR-$10$}: Performance (\%) of the proposed model against various adaptive attacks. MPR denotes the maximum percentage rejected samples.}
\label{table:cifar_adaptive}
\vspace{-0.0cm}
\setlength\tabcolsep{3pt}
\resizebox{1.0\linewidth}{!}{
\begin{tabular}{l|ccccc}
\toprule
& \multicolumn{1}{c}{\textbf{$Acc_{adv, 0\%}\uparrow$}} & \multicolumn{1}{c}{\textbf{$Acc_{adv, 10\%}\uparrow$}} & \multicolumn{1}{c}{\textbf{$FC\uparrow$}} & \multicolumn{1}{c}{\textbf{$FW\downarrow$}}& \multicolumn{1}{c}{MPR} \\
\midrule 

A1: FA (KL, targ)     & 58.50 & 69.71 & 47.20 & 20.50 & 41.90   \\
A2: FA (MSE, targ)    & 59.20 & 71.57 & 48.60 & 19.30 & 43.60 \\
A3: FA (KL, untarg)                 & 62.70 & 74.61 & 50.60 & 17.20 & 39.90 \\
A4: FA (MSE, untarg)                & 62.90 & 74.88 & 49.50 & 16.60 & 41.00 \\
A5: FA (MSE + min var, targ)        & 58.60 & 70.57 & 47.50 & 19.80 & 43.20 \\
A6: Diverse CE (sample)             & 70.20 & 84.27 & 55.20 & 10.30 & 40.30 \\
\midrule 
RA1: Max entropy                    & 66.80 & 84.19 & 45.80 & 8.60  & 50.80 \\
RA2: Max entropy + Min CE           & 77.60 & 88.38 & 62.40 & 8.20  & 33.30 \\
RA3: Output diversify (all classes) & 73.60 & 86.12 & 59.60 & 9.60  & 34.60 \\
RA4: Maximize variance              & 73.80 & 85.67 & 60.40 & 10.10 & 31.60 \\
Ensemble of 6 attacks & 50.40 & 55.39 & 41.10 & 33.10 & 49.50 \\
PGD                   & 53.30 & 67.40  & 42.40 & 20.50 & 43.60 \\
\bottomrule
\end{tabular}}
\end{table}

%% file: cvpr.bbl
\begin{thebibliography}{10}\itemsep=-1pt

\bibitem{andriushchenko2019square}
Maksym Andriushchenko, Francesco Croce, Nicolas Flammarion, and Matthias Hein.
\newblock Square attack: a query-efficient black-box adversarial attack via
  random search.
\newblock {\em arXiv preprint arXiv:1912.00049}, 2019.

\bibitem{athalye2018obfuscated}
Anish Athalye, Nicholas Carlini, and David Wagner.
\newblock Obfuscated gradients give a false sense of security: Circumventing
  defenses to adversarial examples.
\newblock {\em arXiv preprint arXiv:1802.00420}, 2018.

\bibitem{blum2020random}
Avrim Blum, Travis Dick, Naren Manoj, and Hongyang Zhang.
\newblock Random smoothing might be unable to certify $\ell_{\infty} $
  robustness for high-dimensional images.
\newblock {\em arXiv preprint arXiv:2002.03517}, 2020.

\bibitem{buckman2018thermometer}
Jacob Buckman, Aurko Roy, Colin Raffel, and Ian Goodfellow.
\newblock Thermometer encoding: One hot way to resist adversarial examples.
\newblock In {\em International Conference on Learning Representations}, 2018.

\bibitem{carlini2019evaluating}
Nicholas Carlini, Anish Athalye, Nicolas Papernot, Wieland Brendel, Jonas
  Rauber, Dimitris Tsipras, Ian Goodfellow, and Aleksander Madry.
\newblock On evaluating adversarial robustness.
\newblock {\em arXiv preprint arXiv:1902.06705}, 2019.

\bibitem{carlini2017towards}
Nicholas Carlini and David Wagner.
\newblock Towards evaluating the robustness of neural networks.
\newblock In {\em 2017 IEEE Symposium on Security and Privacy (SP)}, pages
  39--57. IEEE, 2017.

\bibitem{carmon2019unlabeled}
Yair Carmon, Aditi Raghunathan, Ludwig Schmidt, John~C Duchi, and Percy~S
  Liang.
\newblock Unlabeled data improves adversarial robustness.
\newblock In {\em Advances in Neural Information Processing Systems}, pages
  11192--11203, 2019.

\bibitem{cohen2019certified}
Jeremy~M Cohen, Elan Rosenfeld, and J~Zico Kolter.
\newblock Certified adversarial robustness via randomized smoothing.
\newblock {\em arXiv preprint arXiv:1902.02918}, 2019.

\bibitem{croce2019minimally}
Francesco Croce and Matthias Hein.
\newblock Minimally distorted adversarial examples with a fast adaptive
  boundary attack.
\newblock {\em arXiv preprint arXiv:1907.02044}, 2019.

\bibitem{croce2020reliable}
Francesco Croce and Matthias Hein.
\newblock Reliable evaluation of adversarial robustness with an ensemble of
  diverse parameter-free attacks.
\newblock {\em arXiv preprint arXiv:2003.01690}, 2020.

\bibitem{s.2018stochastic}
Guneet~S. Dhillon, Kamyar Azizzadenesheli, Jeremy~D. Bernstein, Jean Kossaifi,
  Aran Khanna, Zachary~C. Lipton, and Animashree Anandkumar.
\newblock Stochastic activation pruning for robust adversarial defense.
\newblock In {\em International Conference on Learning Representations}, 2018.

\bibitem{ghosh2019resisting}
Partha Ghosh, Arpan Losalka, and Michael~J Black.
\newblock Resisting adversarial attacks using gaussian mixture variational
  autoencoders.
\newblock In {\em Proceedings of the AAAI Conference on Artificial
  Intelligence}, volume~33, pages 541--548, 2019.

\bibitem{goodfellow2014explaining}
Ian~J Goodfellow, Jonathon Shlens, and Christian Szegedy.
\newblock Explaining and harnessing adversarial examples.
\newblock {\em arXiv preprint arXiv:1412.6572}, 2014.

\bibitem{he2016deep}
Kaiming He, Xiangyu Zhang, Shaoqing Ren, and Jian Sun.
\newblock Deep residual learning for image recognition.
\newblock In {\em Proceedings of the IEEE conference on computer vision and
  pattern recognition}, pages 770--778, 2016.

\bibitem{he2019parametric}
Zhezhi He, Adnan~Siraj Rakin, and Deliang Fan.
\newblock Parametric noise injection: Trainable randomness to improve deep
  neural network robustness against adversarial attack.
\newblock In {\em Proceedings of the IEEE Conference on Computer Vision and
  Pattern Recognition}, pages 588--597, 2019.

\bibitem{hu2019new}
Shengyuan Hu, Tao Yu, Chuan Guo, Wei-Lun Chao, and Kilian~Q Weinberger.
\newblock A new defense against adversarial images: Turning a weakness into a
  strength.
\newblock In {\em Advances in Neural Information Processing Systems}, pages
  1635--1646, 2019.

\bibitem{hung2019rank}
Kenneth Hung, William Fithian, et~al.
\newblock Rank verification for exponential families.
\newblock {\em The Annals of Statistics}, 47(2):758--782, 2019.

\bibitem{kingma2013auto}
Diederik~P Kingma and Max Welling.
\newblock Auto-encoding variational bayes.
\newblock {\em arXiv preprint arXiv:1312.6114}, 2013.

\bibitem{krizhevsky2009learning}
Alex Krizhevsky et~al.
\newblock Learning multiple layers of features from tiny images.
\newblock 2009.

\bibitem{ma2018characterizing}
Xingjun Ma, Bo Li, Yisen Wang, Sarah~M. Erfani, Sudanthi Wijewickrema, Grant
  Schoenebeck, Michael~E. Houle, Dawn Song, and James Bailey.
\newblock Characterizing adversarial subspaces using local intrinsic
  dimensionality.
\newblock In {\em International Conference on Learning Representations}, 2018.

\bibitem{madry-iclr-2018}
Aleksander Madry, Aleksandar Makelov, Ludwig Schmidt, Tsipras Dimitris, and
  Adrian Vladu.
\newblock Towards deep learning models resistant to adversarial attacks.
\newblock In {\em International Conference on Learning Representations (ICLR)},
  2018.

\bibitem{pang2020bag}
Tianyu Pang, Xiao Yang, Yinpeng Dong, Hang Su, and Jun Zhu.
\newblock Bag of tricks for adversarial training.
\newblock {\em arXiv preprint arXiv:2010.00467}, 2020.

\bibitem{pereyra2017regularizing}
Gabriel Pereyra, George Tucker, Jan Chorowski, {\L}ukasz Kaiser, and Geoffrey
  Hinton.
\newblock Regularizing neural networks by penalizing confident output
  distributions.
\newblock {\em arXiv preprint arXiv:1701.06548}, 2017.

\bibitem{rice2020overfitting}
Leslie Rice, Eric Wong, and J~Zico Kolter.
\newblock Overfitting in adversarially robust deep learning.
\newblock {\em arXiv preprint arXiv:2002.11569}, 2020.

\bibitem{roth2019odds}
Kevin Roth, Yannic Kilcher, and Thomas Hofmann.
\newblock The odds are odd: A statistical test for detecting adversarial
  examples.
\newblock {\em arXiv preprint arXiv:1902.04818}, 2019.

\bibitem{sabour2015adversarial}
Sara Sabour, Yanshuai Cao, Fartash Faghri, and David~J Fleet.
\newblock Adversarial manipulation of deep representations.
\newblock {\em arXiv preprint arXiv:1511.05122}, 2015.

\bibitem{schmidt2018adversarially}
Ludwig Schmidt, Shibani Santurkar, Dimitris Tsipras, Kunal Talwar, and
  Aleksander Madry.
\newblock Adversarially robust generalization requires more data.
\newblock In {\em Advances in Neural Information Processing Systems}, pages
  5014--5026, 2018.

\bibitem{song2018pixeldefend}
Yang Song, Taesup Kim, Sebastian Nowozin, Stefano Ermon, and Nate Kushman.
\newblock Pixeldefend: Leveraging generative models to understand and defend
  against adversarial examples.
\newblock In {\em International Conference on Learning Representations}, 2018.

\bibitem{sriramanan2020gama}
Gaurang Sriramanan, Sravanti Addepalli, Arya Baburaj, and R Venkatesh~Babu.
\newblock {Guided Adversarial Attack for Evaluating and Enhancing Adversarial
  Defenses}.
\newblock In {\em Advances in Neural Information Processing Systems (NeurIPS)},
  2020.

\bibitem{srivastava2014dropout}
Nitish Srivastava, Geoffrey Hinton, Alex Krizhevsky, Ilya Sutskever, and Ruslan
  Salakhutdinov.
\newblock Dropout: a simple way to prevent neural networks from overfitting.
\newblock {\em The journal of machine learning research}, 15(1):1929--1958,
  2014.

\bibitem{stutz2019confidence}
David Stutz, Matthias Hein, and Bernt Schiele.
\newblock Confidence-calibrated adversarial training: Generalizing to unseen
  attacks.
\newblock In {\em Proceedings of the International Conference on Machine
  Learning}, 2020.

\bibitem{tramer2020adaptive}
Florian Tramer, Nicholas Carlini, Wieland Brendel, and Aleksander Madry.
\newblock On adaptive attacks to adversarial example defenses.
\newblock {\em arXiv preprint arXiv:2002.08347}, 2020.

\bibitem{wu2020adversarial}
Dongxian Wu, Shu-Tao Xia, and Yisen Wang.
\newblock Adversarial weight perturbation helps robust generalization.
\newblock {\em Advances in Neural Information Processing Systems}, 33, 2020.

\bibitem{xie2018mitigating}
Cihang Xie, Jianyu Wang, Zhishuai Zhang, Zhou Ren, and Alan Yuille.
\newblock Mitigating adversarial effects through randomization.
\newblock In {\em International Conference on Learning Representations}, 2018.

\bibitem{yin2019adversarial}
Xuwang Yin, Soheil Kolouri, and Gustavo~K Rohde.
\newblock Adversarial example detection and classification with asymmetrical
  adversarial training.
\newblock {\em arXiv preprint arXiv:1905.11475}, 2019.

\bibitem{zhang2017mixup}
Hongyi Zhang, Moustapha Cisse, Yann~N Dauphin, and David Lopez-Paz.
\newblock mixup: Beyond empirical risk minimization.
\newblock {\em arXiv preprint arXiv:1710.09412}, 2017.

\bibitem{zhang2019theoretically}
Hongyang Zhang, Yaodong Yu, Jiantao Jiao, Eric~P Xing, Laurent~El Ghaoui, and
  Michael~I Jordan.
\newblock Theoretically principled trade-off between robustness and accuracy.
\newblock {\em arXiv preprint arXiv:1901.08573}, 2019.

\end{thebibliography}
